\documentclass[oneside,letterpaper,10pt,onecolumn]{article}
\usepackage[utf8]{inputenc}

\usepackage[T1]{fontenc}

\usepackage[numbers]{natbib} 

\usepackage{verbatim}
\usepackage[sumlimits,]{amsmath}
\usepackage{amsfonts}
\usepackage{amsthm}
\usepackage{amssymb}
\usepackage{srcltx}

\usepackage{tocloft}

\usepackage{bm}

\usepackage[bb=boondox]{mathalfa}   

\usepackage{array}
\usepackage{graphics,graphicx,psfrag,epsfig}
\usepackage{wrapfig}
\graphicspath{{.} {./figures/}}

\newcolumntype{M}[1]{>{\centering\arraybackslash}m{#1}}

\usepackage[parfill]{parskip}
\usepackage{geometry}

\usepackage{xcolor}
\usepackage{color}
\definecolor{deepblue}{rgb}{0,0,0.5}
\definecolor{deepred}{rgb}{0.6,0,0}
\definecolor{deepgreen}{rgb}{0,0.5,0}
\definecolor{deeporange}{rgb}{0.6,0.25,0}
\definecolor{verylightgray}{rgb}{0.97,0.97,0.97}
\definecolor{bluegray}{rgb}{0.3,0.3,0.6}

\usepackage[
    colorlinks=true,
    linkcolor=black,       urlcolor=bluegray,
    anchorcolor=bluegray,  filecolor=bluegray,
    menucolor=bluegray,    citecolor=bluegray
]{hyperref}

\usepackage{cleveref}
\usepackage{cancel}
\usepackage{ulem}
\newcommand{\stkout}[1]{\ifmmode\text{\sout{\ensuremath{#1}}}\else\sout{#1}\fi}

\DeclareFixedFont{\ttb}{T1}{txtt}{bx}{n}{8} 
\DeclareFixedFont{\ttm}{T1}{txtt}{m}{n}{8}  

\usepackage{listings}
\usepackage{textcomp}  

\lstdefinestyle{pythonstyle}{
  language=Python,
  basicstyle=\footnotesize\ttm,
  keywordstyle=\ttb\color{deepblue},
  commentstyle=\ttm\color{deepgreen},
  stringstyle=\color{deeporange},
  emphstyle=\ttb\color{deepred},
  backgroundcolor=\color{verylightgray},
  emph={__init__, __call__},      
  morekeywords = {self,yield},
  showspaces=false,               
  showstringspaces=false,         
  showtabs=false,                 
  frame=single,                   
  tabsize=2,                      
  captionpos=b,                   
  breaklines=false,               
  breakatwhitespace=false,        
  columns=fullflexible,           
  extendedchars=false,
  keepspaces=true,
  upquote=true,                   
  literate={-}{-}1 {*}{*}1
}

\lstnewenvironment{python}[1][]
{
\lstset{style=pythonstyle}
\lstset{#1}
}
{}

\newcommand\pythoninline[1]{{\lstset{style=pythonstyle}\lstinline!#1!}}

\newcommand{\iid}{\stackrel{\tiny\mathrm{iid}}{\sim}}

\DeclareMathOperator*{\argmin}{\operatorname{\arg\min}}
\DeclareMathOperator*{\argmax}{\operatorname{\arg\max}}

\newcommand{\eye}{\operatorname{I}}

\theoremstyle{plain}            

\theoremstyle{definition}       

\def\vs{{\bm{s}}}

\def\vx{{\bm{x}}}
\def\vy{{\bm{y}}}
\def\vz{{\bm{z}}}

\usepackage[numbers]{natbib}
\usepackage{mathtools}
\DeclarePairedDelimiterX{\infdivx}[2]{(}{)}{%
  #1\;\delimsize\|\;#2%
}
\newcommand{\infdiv}{D_\textrm{KL}\infdivx}

\newcommand\scalemath[2]{\scalebox{#1}{\mbox{\ensuremath{\displaystyle #2}}}}

\title{Understanding Diffusion Models: A Unified Perspective}
\author{\small \textbf{Calvin Luo} \\ \small Google Research, Brain Team \\ {\tt\small calvinluo@google.com}}
\date{\small \today}

\hypersetup{
  pdftitle={Diffusion Models},
  pdfauthor={Calvin Luo},
  pdfsubject={Diffusion Models},
}

\begin{document}
\maketitle

\tableofcontents
\section*{Introduction: Generative Models}
\addcontentsline{toc}{section}{\protect\numberline{}Introduction: Generative Models}%

Given observed samples $\vx$ from a distribution of interest, the goal of a \textbf{generative model} is to learn to \textit{model} its true data distribution $p(\vx)$.  Once learned, we can \textit{generate} new samples from our approximate model at will.  Furthermore, under some formulations, we are able to use the learned model to evaluate the likelihood of observed or sampled data as well.

There are several well-known directions in current literature, that we will only introduce briefly at a high level.  Generative Adversarial Networks (GANs) model the sampling procedure of a complex distribution, which is learned in an adversarial manner.  Another class of generative models, termed "likelihood-based", seeks to learn a model that assigns a high likelihood to the observed data samples.  This includes autoregressive models, normalizing flows, and Variational Autoencoders (VAEs).  Another similar approach is energy-based modeling, in which a distribution is learned as an arbitrarily flexible energy function that is then normalized.  Score-based generative models are highly related; instead of learning to model the energy function itself, they learn the \textit{score} of the energy-based model as a neural network.  In this work we explore and review diffusion models, which as we will demonstrate, have both likelihood-based and score-based interpretations.  We showcase the math behind such models in excruciating detail, with the aim that anyone can follow along and understand what diffusion models are and how they work.

\section*{Background: ELBO, VAE, and Hierarchical VAE}
\addcontentsline{toc}{section}{\protect\numberline{}Background: ELBO, VAE, and Hierarchical VAE}%

For many modalities, we can think of the data we observe as represented or generated by an associated unseen \textit{latent} variable, which we can denote by random variable $\vz$.  The best intuition for expressing this idea is through Plato's \href{https://en.wikipedia.org/wiki/Allegory_of_the_cave}{Allegory of the Cave}.  In the allegory, a group of people are chained inside a cave their entire life and can only see the two-dimensional shadows projected onto a wall in front of them, which are generated by unseen three-dimensional objects passed before a fire.  To such people, everything they observe is actually determined by higher-dimensional abstract concepts that they can never behold.

Analogously, the objects that we encounter in the actual world may also be generated as a function of some higher-level representations; for example, such representations may encapsulate abstract properties such as color, size, shape, and more.  Then, what we observe can be interpreted as a three-dimensional projection or instantiation of such abstract concepts, just as what the cave people observe is actually a two-dimensional projection of three-dimensional objects.  Whereas the cave people can never see (or even fully comprehend) the hidden objects, they can still reason and draw inferences about them; in a similar way, we can approximate latent representations that describe the data we observe.

Whereas Plato’s Allegory illustrates the idea behind latent variables as potentially unobservable representations that determine observations, a caveat of this analogy is that in generative modeling, we generally seek to learn lower-dimensional latent representations rather than higher-dimensional ones.  This is because trying to learn a representation of higher dimension than the observation is a fruitless endeavor without strong priors.  On the other hand, learning lower-dimensional latents can also be seen as a form of compression, and can potentially uncover semantically meaningful structure describing observations.

\subsubsection*{Evidence Lower Bound}
\addcontentsline{toc}{section}{\protect\numberline{}\protect\numberline{}Evidence Lower Bound}%

Mathematically, we can imagine the latent variables and the data we observe as modeled by a joint distribution $p(\vx, \vz)$.  Recall one approach of generative modeling, termed "likelihood-based", is to learn a model to maximize the likelihood $p(\vx)$ of all observed $\vx$.  There are two ways we can manipulate this joint distribution to recover the likelihood of purely our observed data $p(\vx)$; we can explicitly \href{https://en.wikipedia.org/wiki/Marginal_likelihood}{marginalize} out the latent variable $\vz$:
\begin{equation}
\label{eq:1}
p(\vx) = \int p(\vx, \vz)d\vz
\end{equation}
or, we could also appeal to the \href{https://en.wikipedia.org/wiki/Chain_rule_(probability)}{chain rule of probability}:
\begin{equation}
\label{eq:2}
p(\vx) = \frac{p(\vx, \vz)}{p(\vz|\vx)}
\end{equation}
Directly computing and maximizing the likelihood $p(\vx)$ is difficult because it either involves integrating out all latent variables $\vz$ in Equation \ref{eq:1}, which is intractable for complex models, or it involves having access to a ground truth latent encoder $p(\vz|\vx)$ in Equation \ref{eq:2}.  However, using these two equations, we can derive a term called the \textbf{E}vidence \textbf{L}ower \textbf{Bo}und (ELBO), which as its name suggests, is a \href{https://en.wikipedia.org/wiki/Upper_and_lower_bounds}{lower bound} of the evidence.  The evidence is quantified in this case as the log likelihood of the observed data.  Then, maximizing the ELBO becomes a proxy objective with which to optimize a latent variable model; in the best case, when the ELBO is powerfully parameterized and perfectly optimized, it becomes exactly equivalent to the evidence.  Formally, the equation of the ELBO is:
\begin{equation}
\mathbb{E}_{q_{\bm{\phi}}(\vz|\vx)}\left[\log\frac{p(\vx, \vz)}{q_{\bm{\phi}}(\vz|\vx)}\right]
\end{equation}
To make the relationship with the evidence explicit, we can mathematically write:
\begin{equation}
\log p(\vx) \geq \mathbb{E}_{q_{\bm{\phi}}(\vz|\vx)}\left[\log\frac{p(\vx, \vz)}{q_{\bm{\phi}}(\vz|\vx)}\right]
\end{equation}
Here, $q_{\bm{\phi}}(\vz|\vx)$ is a flexible approximate variational distribution with parameters $\bm{\phi}$ that we seek to optimize.  Intuitively, it can be thought of as a parameterizable model that is learned to estimate the true distribution over latent variables for given observations $\vx$; in other words, it seeks to approximate true posterior $p(\vz|\vx)$.  As we will see when exploring the Variational Autoencoder, as we increase the lower bound by tuning the parameters $\bm{\phi}$ to maximize the ELBO, we gain access to components that can be used to model the true data distribution and sample from it, thus learning a generative model.  For now, let us try to dive deeper into why the ELBO is an objective we would like to maximize.

Let us begin by deriving the ELBO, using Equation \ref{eq:1}:
\begin{align}
\log p(\vx) & = \log \int p(\vx, \vz)d\vz && \text{(Apply Equation \ref{eq:1})}\\
           & = \log \int \frac{p(\vx, \vz)q_{\bm{\phi}}(\vz|\vx)}{q_{\bm{\phi}}(\vz|\vx)}d\vz && \text{(Multiply by $1 = \frac{q_{\bm{\phi}}(\vz|\vx)}{q_{\bm{\phi}}(\vz|\vx)}$)}\\
           & = \log \mathbb{E}_{q_{\bm{\phi}}(\vz|\vx)}\left[\frac{p(\vx, \vz)}{q_{\bm{\phi}}(\vz|\vx)}\right] && \text{(Definition of Expectation)}\\
           & \geq \mathbb{E}_{q_{\bm{\phi}}(\vz|\vx)}\left[\log \frac{p(\vx, \vz)}{q_{\bm{\phi}}(\vz|\vx)}\right] && \text{(Apply \href{https://en.wikipedia.org/wiki/Jensen\%27s_inequality}{Jensen's Inequality})} \label{eq:8}
\end{align}
In this derivation, we directly arrive at our lower bound by applying Jensen's Inequality.  However, this does not supply us much useful information about what is actually going on underneath the hood; crucially, this proof gives no intuition on exactly why the ELBO is actually a lower bound of the evidence, as Jensen's Inequality handwaves it away.  Furthermore, simply knowing that the ELBO is truly a lower bound of the data does not really tell us why we want to maximize it as an objective.  To better understand the relationship between the evidence and the ELBO, let us perform another derivation, this time using Equation \ref{eq:2}:
\begin{align}
\log p(\vx) & = \log p(\vx) \int q_{\bm{\phi}}(\vz|\vx)dz && \text{(Multiply by $1 = \int q_{\bm{\phi}}(\vz|\vx)d\vz$)}\\
          & = \int q_{\bm{\phi}}(\vz|\vx)(\log p(\vx))dz && \text{(Bring evidence into integral)}\\
          & = \mathbb{E}_{q_{\bm{\phi}}(\vz|\vx)}\left[\log p(\vx)\right] && \text{(Definition of Expectation)}\\
          & = \mathbb{E}_{q_{\bm{\phi}}(\vz|\vx)}\left[\log\frac{p(\vx, \vz)}{p(\vz|\vx)}\right]&& \text{(Apply Equation \ref{eq:2})}\\
          & = \mathbb{E}_{q_{\bm{\phi}}(\vz|\vx)}\left[\log\frac{p(\vx, \vz)q_{\bm{\phi}}(\vz|\vx)}{p(\vz|\vx)q_{\bm{\phi}}(\vz|\vx)}\right]&& \text{(Multiply by $1 = \frac{q_{\bm{\phi}}(\vz|\vx)}{q_{\bm{\phi}}(\vz|\vx)}$)}\\
          & = \mathbb{E}_{q_{\bm{\phi}}(\vz|\vx)}\left[\log\frac{p(\vx, \vz)}{q_{\bm{\phi}}(\vz|\vx)}\right] + \mathbb{E}_{q_{\bm{\phi}}(\vz|\vx)}\left[\log\frac{q_{\bm{\phi}}(\vz|\vx)}{p(\vz|\vx)}\right] && \text{(Split the Expectation)}\\
          & = \mathbb{E}_{q_{\bm{\phi}}(\vz|\vx)}\left[\log\frac{p(\vx, \vz)}{q_{\bm{\phi}}(\vz|\vx)}\right] + \infdiv{q_{\bm{\phi}}(\vz|\vx)}{p(\vz|\vx)}  && \text{(Definition of \href{https://en.wikipedia.org/wiki/Kullback\%E2\%80\%93Leibler_divergence}{KL Divergence})}\label{eq:15}\\
          & \geq \mathbb{E}_{q_{\bm{\phi}}(\vz|\vx)}\left[\log\frac{p(\vx, \vz)}{q_{\bm{\phi}}(\vz|\vx)}\right]  && \text{(KL Divergence always $\geq 0$)}
\end{align}
From this derivation, we clearly observe from Equation \ref{eq:15} that the evidence is equal to the ELBO plus the KL Divergence between the approximate posterior $q_{\bm{\phi}}(\vz|\vx)$ and the true posterior $p(\vz|\vx)$.  In fact, it was this KL Divergence term that was magically removed by Jensen's Inequality in Equation \ref{eq:8} of the first derivation.  Understanding this term is the key to understanding not only the relationship between the ELBO and the evidence, but also the reason why optimizing the ELBO is an appropriate objective at all.

Firstly, we now know why the ELBO is indeed a lower bound: the difference between the evidence and the ELBO is a strictly non-negative KL term, thus the value of the ELBO can never exceed the evidence.

Secondly, we explore why we seek to maximize the ELBO.  Having introduced latent variables $\vz$ that we would like to model, our goal is to learn this underlying latent structure that describes our observed data.  In other words, we want to optimize the parameters of our variational posterior $q_{\bm{\phi}}(\vz|\vx)$ to exactly match the true posterior distribution $p(\vz|\vx)$, which is achieved by minimizing their KL Divergence (ideally to zero).  Unfortunately, it is intractable to minimize this KL Divergence term directly, as we do not have access to the ground truth $p(\vz|\vx)$ distribution.  However, notice that on the left hand side of Equation \ref{eq:15}, the likelihood of our data (and therefore our evidence term $\log p(\vx)$) is always a constant with respect to $\bm{\phi}$, as it is computed by marginalizing out all latents $\vz$ from the joint distribution $p(\vx, \vz)$ and does not depend on $\bm{\phi}$ whatsoever.  Since the ELBO and KL Divergence terms sum up to a constant, any maximization of the ELBO term with respect to $\bm{\phi}$ necessarily invokes an equal minimization of the KL Divergence term.  Thus, the ELBO can be maximized as a proxy for learning how to perfectly model the true latent posterior distribution; the more we optimize the ELBO, the closer our approximate posterior gets to the true posterior.  Additionally, once trained, the ELBO can be used to estimate the likelihood of observed or generated data as well, since it is learned to approximate the model evidence $\log p(\vx)$.

\subsubsection*{Variational Autoencoders}
\addcontentsline{toc}{section}{\protect\numberline{}\protect\numberline{}Variational Autoencoders}%
\begin{figure}
  \centering
  \includegraphics[width=0.25\linewidth]{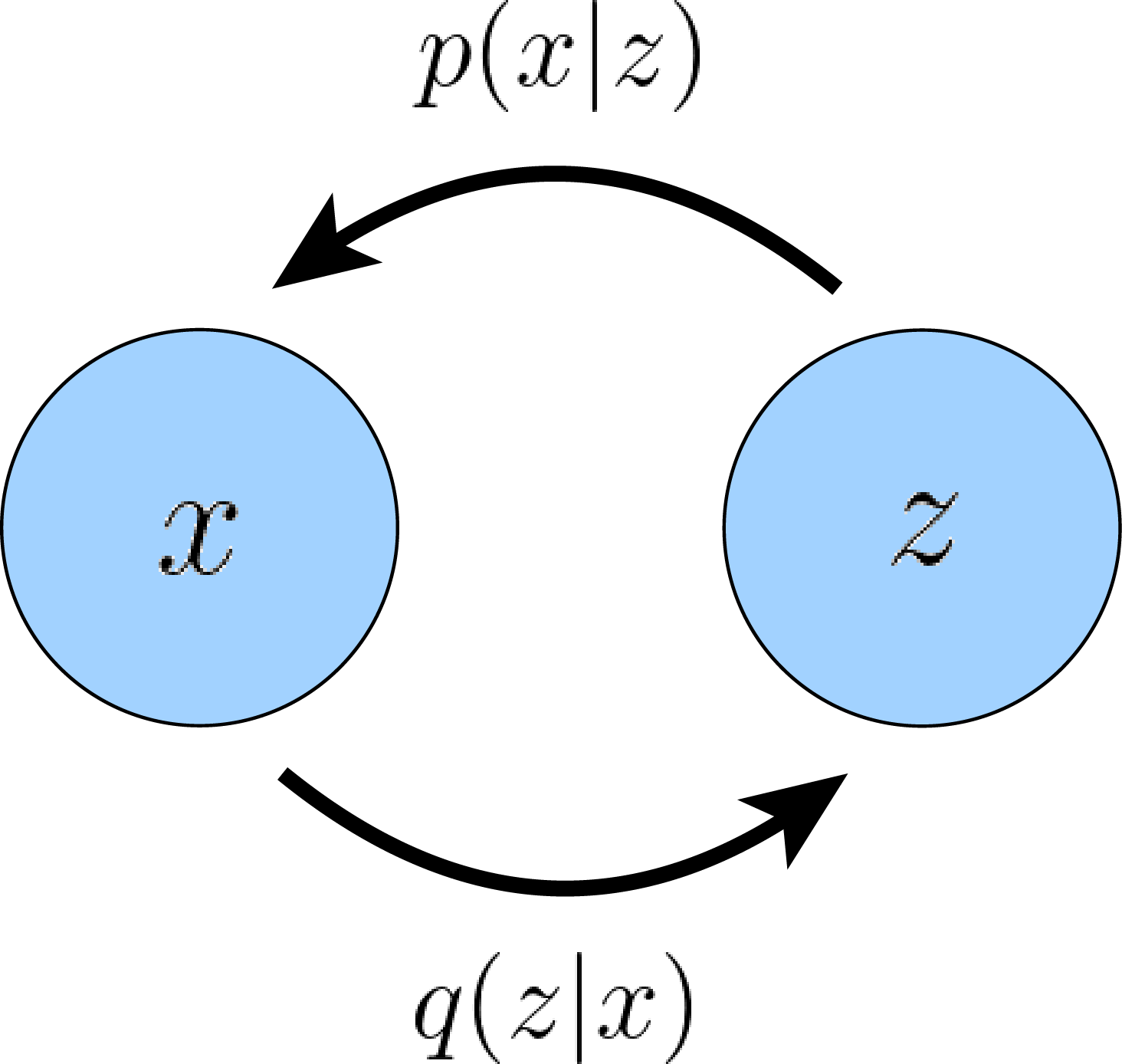}
  \caption{A Variational Autoencoder graphically represented.  Here, encoder $q(\vz|\vx)$ defines a distribution over latent variables $\vz$ for observations $\vx$, and $p(\vx|\vz)$ decodes latent variables into observations.}
  \label{fig:vae}
\end{figure}

In the default formulation of the Variational Autoencoder (VAE)~\cite{kingma2013auto}, we directly maximize the ELBO.  This approach is \textit{variational}, because we optimize for the best $q_{\bm{\phi}}(\vz|\vx)$ amongst a family of potential posterior distributions parameterized by $\bm{\phi}$.  It is called an \textit{autoencoder} because it is reminiscent of a traditional autoencoder model, where input data is trained to predict itself after undergoing an intermediate bottlenecking representation step.  To make this connection explicit, let us dissect the ELBO term further:
\begin{align}
\scalemath{0.98}{\mathbb{E}_{q_{\bm{\phi}}(\vz|\vx)}\left[\log\frac{p(\vx, \vz)}{q_{\bm{\phi}}(\vz|\vx)}\right]}
&= \scalemath{0.98}{\mathbb{E}_{q_{\bm{\phi}}(\vz|\vx)}\left[\log\frac{p_{\bm{\theta}}(\vx|\vz)p(\vz)}{q_{\bm{\phi}}(\vz|\vx)}\right]}         && \scalemath{0.98}{\text{(Chain Rule of Probability)}}\\
&= \scalemath{0.98}{\mathbb{E}_{q_{\bm{\phi}}(\vz|\vx)}\left[\log p_{\bm{\theta}}(\vx|\vz)\right] + \mathbb{E}_{q_{\bm{\phi}}(\vz|\vx)}\left[\log\frac{p(\vz)}{q_{\bm{\phi}}(\vz|\vx)}\right]}         && \scalemath{0.98}{\text{(Split the Expectation)}}\\
&= \underbrace{\scalemath{0.98}{\mathbb{E}_{q_{\bm{\phi}}(\vz|\vx)}\left[\log p_{\bm{\theta}}(\vx|\vz)\right]}}_\text{reconstruction term} - \underbrace{\scalemath{0.98}{\infdiv{q_{\bm{\phi}}(\vz|\vx)}{p(\vz)}}}_\text{prior matching term} && \scalemath{0.98}{\text{(Definition of KL Divergence)}}\label{eq:19}
\end{align}
In this case, we learn an intermediate bottlenecking distribution $q_{\bm{\phi}}(\vz|\vx)$ that can be treated as an \textit{encoder}; it transforms inputs into a distribution over possible latents.  Simultaneously, we learn a deterministic function $p_{\bm{\theta}}(\vx|\vz)$ to convert a given latent vector $\vz$ into an observation $\vx$, which can be interpreted as a \textit{decoder}.

The two terms in Equation \ref{eq:19} each have intuitive descriptions: the first term measures the reconstruction likelihood of the decoder from our variational distribution; this ensures that the learned distribution is modeling effective latents that the original data can be regenerated from.  The second term measures how similar the learned variational distribution is to a prior belief held over latent variables.  Minimizing this term encourages the encoder to actually learn a distribution rather than collapse into a Dirac delta function.  Maximizing the ELBO is thus equivalent to maximizing its first term and minimizing its second term.

A defining feature of the VAE is how the ELBO is optimized jointly over parameters $\bm{\phi}$ and $\bm{\theta}$.  The encoder of the VAE is commonly chosen to model a multivariate Gaussian with diagonal covariance, and the prior is often selected to be a standard multivariate Gaussian: 
\begin{align}
    q_{\bm{\phi}}(\vz|\vx) &= \mathcal{N}(\vz; \bm{\mu}_{\bm{\phi}}(\vx), \bm{\sigma}_{\bm{\phi}}^2(\vx)\textbf{I})\\
    p(\vz) &= \mathcal{N}(\vz; \bm{0}, \textbf{I})
\end{align}
Then, the KL divergence term of the ELBO can be computed analytically, and the reconstruction term can be approximated using a Monte Carlo estimate.  Our objective can then be rewritten as:
\begin{align}
    \scalemath{0.97}{\argmax_{\bm{\phi}, \bm{\theta}} \mathbb{E}_{q_{\bm{\phi}}(\vz|\vx)}\left[\log p_{\bm{\theta}}(\vx|\vz)\right] - \infdiv{q_{\bm{\phi}}(\vz|\vx)}{p(\vz)} \approx \argmax_{\bm{\phi}, \bm{\theta}} \sum_{l=1}^{L}\log p_{\bm{\theta}}(\vx|\vz^{(l)}) - \infdiv{q_{\bm{\phi}}(\vz|\vx)}{p(\vz)}}
\end{align}
where latents $\{\vz^{(l)}\}_{l=1}^L$ are sampled from $q_{\bm{\phi}}(\vz|\vx)$, for every observation $\vx$ in the dataset.  However, a problem arises in this default setup: each $\vz^{(l)}$ that our loss is computed on is generated by a stochastic sampling procedure, which is generally non-differentiable.  Fortunately, this can be addressed via the \textit{reparameterization trick} when $q_{\bm{\phi}}(\vz|\vx)$ is designed to model certain distributions, including the multivariate Gaussian.

The reparameterization trick rewrites a random variable as a deterministic function of a noise variable; this allows for the optimization of the non-stochastic terms through gradient descent.  For example, samples from a normal distribution $x \sim \mathcal{N}(x;\mu, \sigma^2)$ with arbitrary mean $\mu$ and variance $\sigma^2$ can be rewritten as:
\begin{align*}
    x &= \mu + \sigma\epsilon \quad \text{with } \epsilon \sim \mathcal{N}(\epsilon; 0, \eye)
\end{align*}
In other words, arbitrary Gaussian distributions can be interpreted as standard Gaussians (of which $\epsilon$ is a sample) that have their mean shifted from zero to the target mean $\mu$ by addition, and their variance stretched by the target variance $\sigma^2$.  Therefore, by the reparameterization trick, sampling from an arbitrary Gaussian distribution can be performed by sampling from a standard Gaussian, scaling the result by the target standard deviation, and shifting it by the target mean.

In a VAE, each $\vz$ is thus computed as a deterministic function of input $\vx$ and auxiliary noise variable $\bm{\epsilon}$:
\begin{align*}
    \vz &= \bm{\mu}_{\bm{\phi}}(\vx) + \bm{\sigma}_{\bm{\phi}}(\vx)\odot\bm{\epsilon} \quad \text{with } \bm{\epsilon} \sim \mathcal{N}(\bm{\epsilon};\bm{0}, \textbf{I})
\end{align*}
where $\odot$ represents an element-wise product.  Under this reparameterized version of $\vz$, gradients can then be computed with respect to $\bm{\phi}$ as desired, to optimize $\bm{\mu}_{\bm{\phi}}$ and $\bm{\sigma}_{\bm{\phi}}$.  The VAE therefore utilizes the reparameterization trick and Monte Carlo estimates to optimize the ELBO jointly over $\bm{\phi}$ and $\bm{\theta}$.

After training a VAE, generating new data can be performed by sampling directly from the latent space $p(\vz)$ and then running it through the decoder.  Variational Autoencoders are particularly interesting when the dimensionality of $\vz$ is less than that of input $\vx$, as we might then be learning compact, useful representations.  Furthermore, when a semantically meaningful latent space is learned, latent vectors can be edited before being passed to the decoder to more precisely control the data generated.

\subsubsection*{Hierarchical Variational Autoencoders}
\addcontentsline{toc}{section}{\protect\numberline{}\protect\numberline{}Hierarchical Variational Autoencoders}%
A Hierarchical Variational Autoencoder (HVAE)~\cite{kingma2016improved, sonderby2016ladder} is a generalization of a VAE that extends to multiple hierarchies over latent variables.  Under this formulation, latent variables themselves are interpreted as generated from other higher-level, more abstract latents. Intuitively, just as we treat our three-dimensional observed objects as generated from a higher-level abstract latent, the people in Plato's cave treat three-dimensional objects as latents that generate their two-dimensional observations.  Therefore, from the perspective of Plato's cave dwellers, their observations can be treated as modeled by a latent hierarchy of depth two (or more).

\begin{figure}
  \centering
  \includegraphics[width=0.6\linewidth]{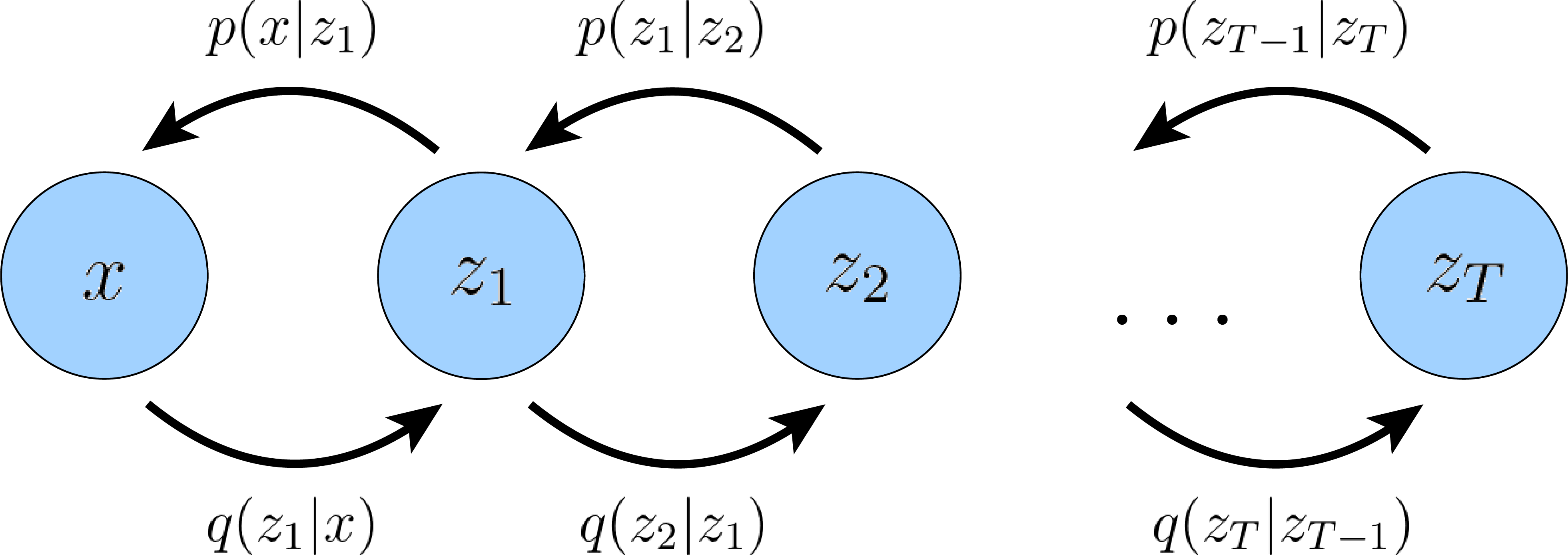}
  \caption{A Markovian Hierarchical Variational Autoencoder with $T$ hierarchical latents.  The generative process is modeled as a Markov chain, where each latent $\vz_t$ is generated only from the previous latent $\vz_{t+1}$.}
  \label{fig:hvae}
\end{figure}

Whereas in the general HVAE with $T$ hierarchical levels, each latent is allowed to condition on all previous latents, in this work we focus on a special case which we call a Markovian HVAE (MHVAE).  In a MHVAE, the generative process is a Markov chain; that is, each transition down the hierarchy is Markovian, where decoding each latent $\vz_t$ only conditions on previous latent $\vz_{t+1}$.  Intuitively, and visually, this can be seen as simply stacking VAEs on top of each other, as depicted in Figure \ref{fig:hvae}; another appropriate term describing this model is a Recursive VAE.  Mathematically, we represent the joint distribution and the posterior of a Markovian HVAE as:
\begin{align}
    p(\vx, \vz_{1:T}) &= p(\vz_T)p_{\bm{\theta}}(\vx|\vz_1)\prod_{t=2}^{T}p_{\bm{\theta}}(\vz_{t-1}|\vz_{t}) \label{eq:20}\\
    q_{\bm{\phi}}(\vz_{1:T}|\vx) &= q_{\bm{\phi}}(\vz_1|\vx)\prod_{t=2}^{T}q_{\bm{\phi}}(\vz_{t}|\vz_{t-1}) \label{eq:21}
\end{align}
Then, we can easily extend the ELBO to be:
\begin{align}
\log p(\vx) &= \log \int p(\vx, \vz_{1:T}) d\vz_{1:T}         && \text{(Apply Equation \ref{eq:1})}\\
&= \log \int \frac{p(\vx, \vz_{1:T})q_{\bm{\phi}}(\vz_{1:T}|\vx)}{q_{\bm{\phi}}(\vz_{1:T}|\vx)} d\vz_{1:T}         && \text{(Multiply by 1 = $\frac{q_{\bm{\phi}}(\vz_{1:T}|\vx)}{q_{\bm{\phi}}(\vz_{1:T}|\vx)}$)}\\
&= \log \mathbb{E}_{q_{\bm{\phi}}(\vz_{1:T}|\vx)}\left[\frac{p(\vx, \vz_{1:T})}{q_{\bm{\phi}}(\vz_{1:T}|\vx)}\right]         && \text{(Definition of Expectation)}\\
&\geq \mathbb{E}_{q_{\bm{\phi}}(\vz_{1:T}|\vx)}\left[\log \frac{p(\vx, \vz_{1:T})}{q_{\bm{\phi}}(\vz_{1:T}|\vx)}\right]         && \text{(Apply Jensen's Inequality)} \label{eq:25}
\end{align}
We can then plug our joint distribution (Equation \ref{eq:20}) and posterior (Equation \ref{eq:21}) into Equation \ref{eq:25} to produce an alternate form:
\begin{align}
\mathbb{E}_{q_{\bm{\phi}}(\vz_{1:T}|\vx)}\left[\log \frac{p(\vx, \vz_{1:T})}{q_{\bm{\phi}}(\vz_{1:T}|\vx)}\right]
&= \mathbb{E}_{q_{\bm{\phi}}(\vz_{1:T}|\vx)}\left[\log \frac{p(\vz_T)p_{\bm{\theta}}(\vx|\vz_1)\prod_{t=2}^{T}p_{\bm{\theta}}(\vz_{t-1}|\vz_{t})}{q_{\bm{\phi}}(\vz_1|\vx)\prod_{t=2}^{T}q_{\bm{\phi}}(\vz_{t}|\vz_{t-1})}\right]
\end{align}
As we will show below, when we investigate Variational Diffusion Models, this objective can be further decomposed into interpretable components.
\begin{figure}
  \centering
  \includegraphics[width=\linewidth]{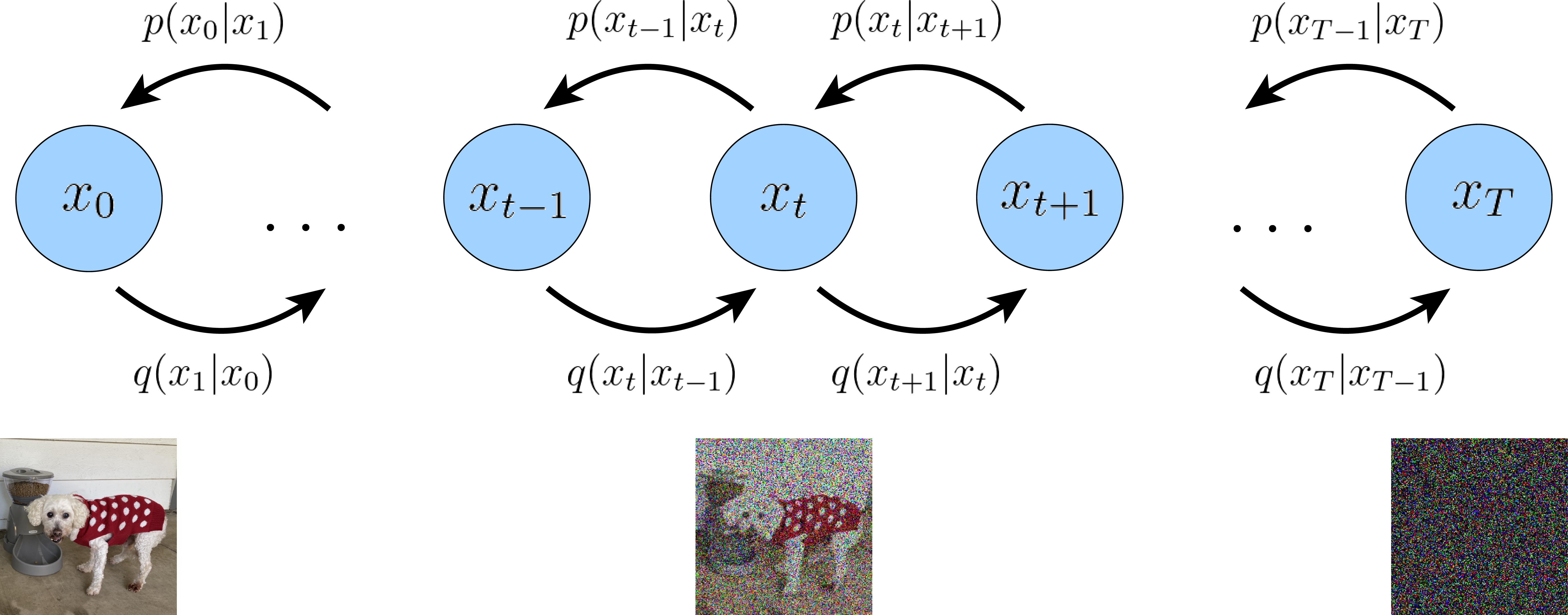}
  \caption{A visual representation of a Variational Diffusion Model; $\vx_0$ represents true data observations such as natural images, $\vx_T$ represents pure Gaussian noise, and $\vx_t$ is an intermediate noisy version of $\vx_0$.  Each $q(\vx_t|\vx_{t-1})$ is modeled as a Gaussian distribution that uses the output of the previous state as its mean.}
  \label{fig:vdm}
\end{figure}
\section*{Variational Diffusion Models}
\addcontentsline{toc}{section}{\protect\numberline{}Variational Diffusion Models}%
The easiest way to think of a Variational Diffusion Model (VDM)~\cite{sohl2015deep, ho2020denoising, kingma2021variational} is simply as a Markovian Hierarchical Variational Autoencoder with three key restrictions:
\begin{itemize}
    \item The latent dimension is exactly equal to the data dimension
    \item The structure of the latent encoder at each timestep is not learned; it is pre-defined as a linear Gaussian model.  In other words, it is a Gaussian distribution centered around the output of the previous timestep
    \item The Gaussian parameters of the latent encoders vary over time in such a way that the distribution of the latent at final timestep $T$ is a standard Gaussian
\end{itemize}
Furthermore, we explicitly maintain the Markov property between hierarchical transitions from a standard Markovian Hierarchical Variational Autoencoder.

Let us expand on the implications of these assumptions.  From the first restriction, with some abuse of notation, we can now represent both true data samples and latent variables as $\vx_t$, where $t=0$ represents true data samples and $t \in \left[1, T\right]$ represents a corresponding latent with hierarchy indexed by $t$.  The VDM posterior is the same as the MHVAE posterior (Equation \ref{eq:21}), but can now be rewritten as:
\begin{align}
    q(\vx_{1:T}|\vx_0) = \prod_{t = 1}^{T}q(\vx_{t}|\vx_{t-1})
\end{align}
From the second assumption, we know that the distribution of each latent variable in the encoder is a Gaussian centered around its previous hierarchical latent.  Unlike a Markovian HVAE, the structure of the encoder at each timestep $t$ is not learned; it is fixed as a linear Gaussian model, where the mean and standard deviation can be set beforehand as hyperparameters~\cite{ho2020denoising}, or learned as parameters~\cite{kingma2021variational}.  We parameterize the Gaussian encoder with mean $\bm{\mu}_t(\vx_t) = \sqrt{\alpha_t} \vx_{t-1}$, and variance $\bm{\Sigma}_t(\vx_t) = (1 - \alpha_t) \textbf{I}$, where the form of the coefficients are chosen such that the variance of the latent variables stay at a similar scale; in other words, the encoding process is \textit{variance-preserving}.  Note that alternate Gaussian parameterizations are allowed, and lead to similar derivations.  The main takeaway is that $\alpha_t$ is a (potentially learnable) coefficient that can vary with the hierarchical depth $t$, for flexibility.  Mathematically, encoder transitions are denoted as:
\begin{align}
    q(\vx_{t}|\vx_{t-1}) = \mathcal{N}(\vx_{t} ; \sqrt{\alpha_t} \vx_{t-1}, (1 - \alpha_t) \textbf{I}) \label{eq:27}
\end{align}
From the third assumption, we know that $\alpha_t$ evolves over time according to a fixed or learnable schedule structured such that the distribution of the final latent $p(\vx_T)$ is a standard Gaussian.  We can then update the joint distribution of a Markovian HVAE (Equation \ref{eq:20}) to write the joint distribution for a VDM as: 
\begin{align}
p(\vx_{0:T}) &= p(\vx_T)\prod_{t=1}^{T}p_{\bm{\theta}}(\vx_{t-1}|\vx_t) \label{eq:36} \\
\text{where,}&\nonumber\\
p(\vx_T) &= \mathcal{N}(\vx_T; \bm{0}, \textbf{I})
\end{align}
Collectively, what this set of assumptions describes is a steady noisification of an image input over time; we progressively corrupt an image by adding Gaussian noise until eventually it becomes completely identical to pure Gaussian noise.  Visually, this process is depicted in Figure \ref{fig:vdm}.

Note that our encoder distributions $q(\vx_t|\vx_{t-1})$ are no longer parameterized by $\bm{\phi}$, as they are completely modeled as Gaussians with defined mean and variance parameters at each timestep.  Therefore, in a VDM, we are only interested in learning conditionals $p_{\bm{\theta}}(\vx_{t-1}|\vx_{t})$, so that we can simulate new data.  After optimizing the VDM, the sampling procedure is as simple as sampling Gaussian noise from $p(\vx_T)$ and iteratively running the denoising transitions $p_{\bm{\theta}}(\vx_{t-1}|\vx_{t})$ for $T$ steps to generate a novel $\vx_0$.

Like any HVAE, the VDM can be optimized by maximizing the ELBO, which can be derived as:
\begingroup
\allowdisplaybreaks
\begin{align}
\scalemath{0.9}{\log p(\vx)}
&= \scalemath{0.9}{\log \int p(\vx_{0:T}) d\vx_{1:T}}\\
&= \scalemath{0.9}{\log \int \frac{p(\vx_{0:T})q(\vx_{1:T}|\vx_0)}{q(\vx_{1:T}|\vx_0)} d\vx_{1:T}}\\
&= \scalemath{0.9}{\log \mathbb{E}_{q(\vx_{1:T}|\vx_0)}\left[\frac{p(\vx_{0:T})}{q(\vx_{1:T}|\vx_0)}\right]}\\
&\geq \scalemath{0.9}{\mathbb{E}_{q(\vx_{1:T}|\vx_0)}\left[\log \frac{p(\vx_{0:T})}{q(\vx_{1:T}|\vx_0)}\right]} \label{eq:34}\\
&= \scalemath{0.9}{\mathbb{E}_{q(\vx_{1:T}|\vx_0)}\left[\log \frac{p(\vx_T)\prod_{t=1}^{T}p_{\bm{\theta}}(\vx_{t-1}|\vx_t)}{\prod_{t = 1}^{T}q(\vx_{t}|\vx_{t-1})}\right]}\\
&= \scalemath{0.9}{\mathbb{E}_{q(\vx_{1:T}|\vx_0)}\left[\log \frac{p(\vx_T)p_{\bm{\theta}}(\vx_0|\vx_1)\prod_{t=2}^{T}p_{\bm{\theta}}(\vx_{t-1}|\vx_t)}{q(\vx_T|\vx_{T-1})\prod_{t = 1}^{T-1}q(\vx_{t}|\vx_{t-1})}\right]}\\
&= \scalemath{0.9}{\mathbb{E}_{q(\vx_{1:T}|\vx_0)}\left[\log \frac{p(\vx_T)p_{\bm{\theta}}(\vx_0|\vx_1)\prod_{t=1}^{T-1}p_{\bm{\theta}}(\vx_{t}|\vx_{t+1})}{q(\vx_T|\vx_{T-1})\prod_{t = 1}^{T-1}q(\vx_{t}|\vx_{t-1})}\right]}\\
&= \scalemath{0.9}{\mathbb{E}_{q(\vx_{1:T}|\vx_0)}\left[\log \frac{p(\vx_T)p_{\bm{\theta}}(\vx_0|\vx_1)}{q(\vx_T|\vx_{T-1})}\right] + \mathbb{E}_{q(\vx_{1:T}|\vx_0)}\left[\log \prod_{t = 1}^{T-1}\frac{p_{\bm{\theta}}(\vx_{t}|\vx_{t+1})}{q(\vx_{t}|\vx_{t-1})}\right]}\\
&= \scalemath{0.9}{\mathbb{E}_{q(\vx_{1:T}|\vx_0)}\left[\log p_{\bm{\theta}}(\vx_0|\vx_1)\right] + \mathbb{E}_{q(\vx_{1:T}|\vx_0)}\left[\log \frac{p(\vx_T)}{q(\vx_T|\vx_{T-1})}\right] + \mathbb{E}_{q(\vx_{1:T}|\vx_0)}\left[ \sum_{t=1}^{T-1} \log \frac{p_{\bm{\theta}}(\vx_{t}|\vx_{t+1})}{q(\vx_{t}|\vx_{t-1})}\right]}\\
&= \scalemath{0.9}{\mathbb{E}_{q(\vx_{1:T}|\vx_0)}\left[\log p_{\bm{\theta}}(\vx_0|\vx_1)\right] + \mathbb{E}_{q(\vx_{1:T}|\vx_0)}\left[\log \frac{p(\vx_T)}{q(\vx_T|\vx_{T-1})}\right] + \sum_{t=1}^{T-1}\mathbb{E}_{q(\vx_{1:T}|\vx_0)}\left[ \log \frac{p_{\bm{\theta}}(\vx_{t}|\vx_{t+1})}{q(\vx_{t}|\vx_{t-1})}\right]}\\
&= \scalemath{0.9}{\mathbb{E}_{q(\vx_{1}|\vx_0)}\left[\log p_{\bm{\theta}}(\vx_0|\vx_1)\right] + \mathbb{E}_{q(\vx_{T-1}, \vx_T|\vx_0)}\left[\log \frac{p(\vx_T)}{q(\vx_T|\vx_{T-1})}\right] + \sum_{t=1}^{T-1}\mathbb{E}_{q(\vx_{t-1}, \vx_t, \vx_{t+1}|\vx_0)}\left[\log \frac{p_{\bm{\theta}}(\vx_{t}|\vx_{t+1})}{q(\vx_{t}|\vx_{t-1})}\right]}\\
&=  \begin{aligned}[t]
      \scalemath{0.9}{\underbrace{\mathbb{E}_{q(\vx_{1}|\vx_0)}\left[\log p_{\theta}(\vx_0|\vx_1)\right]}_\text{reconstruction term}} &- \scalemath{0.9}{\underbrace{\mathbb{E}_{q(\vx_{T-1}|\vx_0)}\left[\infdiv{q(\vx_T|\vx_{T-1})}{p(\vx_T)}\right]}_\text{prior matching term}} \\
      &- \scalemath{0.9}{\sum_{t=1}^{T-1}\underbrace{\mathbb{E}_{q(\vx_{t-1}, \vx_{t+1}|\vx_0)}\left[\infdiv{q(\vx_{t}|\vx_{t-1})}{p_{\theta}(\vx_{t}|\vx_{t+1})}\right]}_\text{consistency term}}
    \end{aligned} \label{eq:45}
\end{align}
\endgroup
The derived form of the ELBO can be interpreted in terms of its individual components:
\begin{figure}
  \centering
  \includegraphics[width=\linewidth]{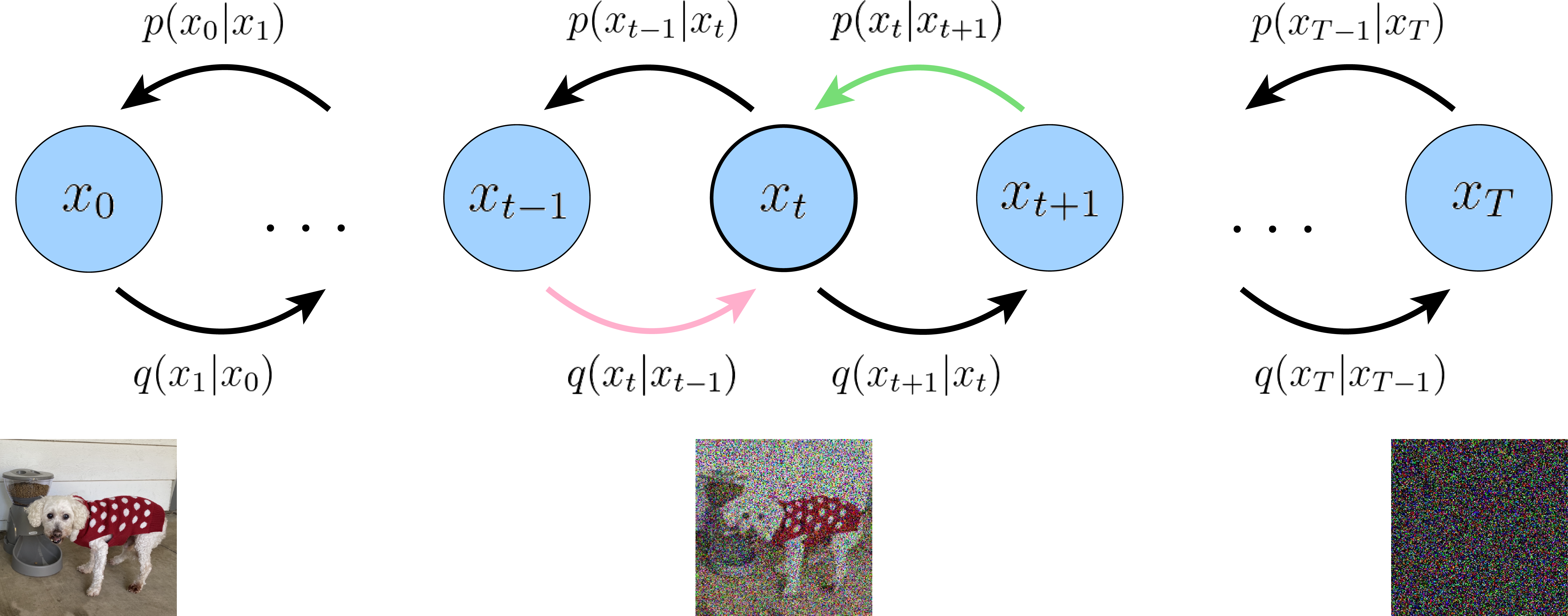}
  \caption{Under our first derivation, a VDM can be optimized by ensuring that for every intermediate $\vx_t$, the posterior from the latent above it $p_{\bm{\theta}}(\vx_t|\vx_{t+1})$ matches the Gaussian corruption of the latent before it $q(\vx_t|\vx_{t-1})$.  In this figure, for each intermediate $\vx_t$, we minimize the difference between the distributions represented by the pink and green arrows.}
  \label{fig:first_deriv}
\end{figure}
\begin{enumerate}
    \item $\mathbb{E}_{q(\vx_{1}|\vx_0)}\left[\log p_{\bm{\theta}}(\vx_0|\vx_1)\right]$ can be interpreted as a \textit{reconstruction term}, predicting the log probability of the original data sample given the first-step latent.  This term also appears in a vanilla VAE, and can be trained similarly.
    \item $\mathbb{E}_{q(\vx_{T-1}|\vx_0)}\left[\infdiv{q(\vx_T|\vx_{T-1})}{p(\vx_T)}\right]$ is a \textit{prior matching term}; it is minimized when the final latent distribution matches the Gaussian prior.  This term requires no optimization, as it has no trainable parameters; furthermore, as we have assumed a large enough $T$ such that the final distribution is Gaussian, this term effectively becomes zero.
    \item $\mathbb{E}_{q(\vx_{t-1}, \vx_{t+1}|\vx_0)}\left[\infdiv{q(\vx_{t}|\vx_{t-1})}{p_{\bm{\theta}}(\vx_{t}|\vx_{t+1})}\right]$ is a \textit{consistency term}; it endeavors to make the distribution at $\vx_t$ consistent, from both forward and backward processes.  That is, a denoising step from a noisier image should match the corresponding noising step from a cleaner image, for every intermediate timestep; this is reflected mathematically by the KL Divergence.  This term is minimized when we train $p_{\theta}(\vx_t|\vx_{t+1})$ to match the Gaussian distribution $q(\vx_t|\vx_{t-1})$, which is defined in Equation \ref{eq:27}.
\end{enumerate}
Visually, this interpretation of the ELBO is depicted in Figure \ref{fig:first_deriv}.  The cost of optimizing a VDM is primarily dominated by the third term, since we must optimize over all timesteps $t$.

Under this derivation, all terms of the ELBO are computed as expectations, and can therefore be approximated using Monte Carlo estimates.  However, actually optimizing the ELBO using the terms we just derived might be suboptimal; because the consistency term is computed as an expectation over two random variables $\left\{\vx_{t-1}, \vx_{t+1}\right\}$ for every timestep, the variance of its Monte Carlo estimate could potentially be higher than a term that is estimated using only one random variable per timestep.  As it is computed by summing up $T-1$ consistency terms, the final estimated value of the ELBO may have high variance for large $T$ values.

Let us instead try to derive a form for our ELBO where each term is computed as an expectation over only one random variable at a time.  The key insight is that we can rewrite encoder transitions as $q(\vx_t|\vx_{t-1}) = q(\vx_t|\vx_{t-1}, \vx_0)$, where the extra conditioning term is superfluous due to the Markov property.  Then, according to Bayes rule, we can rewrite each transition as: 
\begin{align}
q(\vx_t | \vx_{t-1}, \vx_0) = \frac{q(\vx_{t-1}|\vx_t, \vx_0)q(\vx_t|\vx_0)}{q(\vx_{t-1}|\vx_0)}
\end{align}
Armed with this new equation, we can retry the derivation resuming from the ELBO in Equation \ref{eq:34}:
\begingroup
\allowdisplaybreaks
\begin{align}
\scalemath{0.90}{\log p(\vx)}
&\geq \scalemath{0.90}{\mathbb{E}_{q(\vx_{1:T}|\vx_0)}\left[\log \frac{p(\vx_{0:T})}{q(\vx_{1:T}|\vx_0)}\right]}\\
&= \scalemath{0.90}{\mathbb{E}_{q(\vx_{1:T}|\vx_0)}\left[\log \frac{p(\vx_T)\prod_{t=1}^{T}p_{\bm{\theta}}(\vx_{t-1}|\vx_t)}{\prod_{t = 1}^{T}q(\vx_{t}|\vx_{t-1})}\right]}\\
&= \scalemath{0.90}{\mathbb{E}_{q(\vx_{1:T}|\vx_0)}\left[\log \frac{p(\vx_T)p_{\bm{\theta}}(\vx_0|\vx_1)\prod_{t=2}^{T}p_{\bm{\theta}}(\vx_{t-1}|\vx_t)}{q(\vx_1|\vx_0)\prod_{t = 2}^{T}q(\vx_{t}|\vx_{t-1})}\right]}\\
&= \scalemath{0.90}{\mathbb{E}_{q(\vx_{1:T}|\vx_0)}\left[\log \frac{p(\vx_T)p_{\bm{\theta}}(\vx_0|\vx_1)\prod_{t=2}^{T}p_{\bm{\theta}}(\vx_{t-1}|\vx_t)}{q(\vx_1|\vx_0)\prod_{t = 2}^{T}q(\vx_{t}|\vx_{t-1}, \vx_0)}\right]}\\
&= \scalemath{0.90}{\mathbb{E}_{q(\vx_{1:T}|\vx_0)}\left[\log \frac{p_{\bm{\theta}}(\vx_T)p_{\bm{\theta}}(\vx_0|\vx_1)}{q(\vx_1|\vx_0)} + \log \prod_{t=2}^{T}\frac{p_{\bm{\theta}}(\vx_{t-1}|\vx_t)}{q(\vx_{t}|\vx_{t-1}, \vx_0)}\right]}\\
&= \scalemath{0.90}{\mathbb{E}_{q(\vx_{1:T}|\vx_0)}\left[\log \frac{p(\vx_T)p_{\bm{\theta}}(\vx_0|\vx_1)}{q(\vx_1|\vx_0)} + \log \prod_{t=2}^{T}\frac{p_{\bm{\theta}}(\vx_{t-1}|\vx_t)}{\frac{q(\vx_{t-1}|\vx_{t}, \vx_0)q(\vx_t|\vx_0)}{q(\vx_{t-1}|\vx_0)}}\right]}\\
&= \scalemath{0.90}{\mathbb{E}_{q(\vx_{1:T}|\vx_0)}\left[\log \frac{p(\vx_T)p_{\bm{\theta}}(\vx_0|\vx_1)}{q(\vx_1|\vx_0)} + \log \prod_{t=2}^{T}\frac{p_{\bm{\theta}}(\vx_{t-1}|\vx_t)}{\frac{q(\vx_{t-1}|\vx_{t}, \vx_0)\cancel{q(\vx_t|\vx_0)}}{\cancel{q(\vx_{t-1}|\vx_0)}}}\right]}\\
&= \scalemath{0.90}{\mathbb{E}_{q(\vx_{1:T}|\vx_0)}\left[\log \frac{p(\vx_T)p_{\bm{\theta}}(\vx_0|\vx_1)}{\cancel{q(\vx_1|\vx_0)}} + \log \frac{\cancel{q(\vx_1|\vx_0)}}{q(\vx_T|\vx_0)} + \log \prod_{t=2}^{T}\frac{p_{\bm{\theta}}(\vx_{t-1}|\vx_t)}{q(\vx_{t-1}|\vx_{t}, \vx_0)}\right]}\\
&= \scalemath{0.90}{\mathbb{E}_{q(\vx_{1:T}|\vx_0)}\left[\log \frac{p(\vx_T)p_{\bm{\theta}}(\vx_0|\vx_1)}{q(\vx_T|\vx_0)} +  \sum_{t=2}^{T}\log\frac{p_{\bm{\theta}}(\vx_{t-1}|\vx_t)}{q(\vx_{t-1}|\vx_{t}, \vx_0)}\right]}\\
&= \scalemath{0.90}{\mathbb{E}_{q(\vx_{1:T}|\vx_0)}\left[\log p_{\bm{\theta}}(\vx_0|\vx_1)\right] + \mathbb{E}_{q(\vx_{1:T}|\vx_0)}\left[\log \frac{p(\vx_T)}{q(\vx_T|\vx_0)}\right] + \sum_{t=2}^{T}\mathbb{E}_{q(\vx_{1:T}|\vx_0)}\left[\log\frac{p_{\bm{\theta}}(\vx_{t-1}|\vx_t)}{q(\vx_{t-1}|\vx_{t}, \vx_0)}\right]}\\
&= \scalemath{0.90}{\mathbb{E}_{q(\vx_{1}|\vx_0)}\left[\log p_{\bm{\theta}}(\vx_0|\vx_1)\right] + \mathbb{E}_{q(\vx_{T}|\vx_0)}\left[\log \frac{p(\vx_T)}{q(\vx_T|\vx_0)}\right] + \sum_{t=2}^{T}\mathbb{E}_{q(\vx_{t}, \vx_{t-1}|\vx_0)}\left[\log\frac{p_{\bm{\theta}}(\vx_{t-1}|\vx_t)}{q(\vx_{t-1}|\vx_{t}, \vx_0)}\right]}\\
&= \scalemath{0.9}{\underbrace{\mathbb{E}_{q(\vx_{1}|\vx_0)}\left[\log p_{\bm{\theta}}(\vx_0|\vx_1)\right]}_\text{reconstruction term} - \underbrace{\infdiv{q(\vx_T|\vx_0)}{p(\vx_T)}}_\text{prior matching term} - \sum_{t=2}^{T} \underbrace{\mathbb{E}_{q(\vx_{t}|\vx_0)}\left[\infdiv{q(\vx_{t-1}|\vx_t, \vx_0)}{p_{\bm{\theta}}(\vx_{t-1}|\vx_t)}\right]}_\text{denoising matching term}} \label{eq:51}
\end{align}
\endgroup
We have therefore successfully derived an interpretation for the ELBO that can be estimated with lower variance, as each term is computed as an expectation of at most one random variable at a time.  This formulation also has an elegant interpretation, which is revealed when inspecting each individual term:
\begin{enumerate}
    \item $\mathbb{E}_{q(\vx_{1}|\vx_0)}\left[\log p_{\bm{\theta}}(\vx_0|\vx_1)\right]$ can be interpreted as a reconstruction term; like its analogue in the ELBO of a vanilla VAE, this term can be approximated and optimized using a Monte Carlo estimate.
    \item $\infdiv{q(\vx_T|\vx_0)}{p(\vx_T)}$ represents how close the distribution of the final noisified input is to the standard Gaussian prior.  It has no trainable parameters, and is also equal to zero under our assumptions.
    \item $\mathbb{E}_{q(\vx_{t}|\vx_0)}\left[\infdiv{q(\vx_{t-1}|\vx_t, \vx_0)}{p_{\bm{\theta}}(\vx_{t-1}|\vx_t)}\right]$ is a \textit{denoising matching term}.  We learn desired denoising transition step $p_{\bm{\theta}}(\vx_{t-1}|\vx_t)$ as an approximation to tractable, ground-truth denoising transition step $q(\vx_{t-1}|\vx_{t}, \vx_0)$.  The $q(\vx_{t-1}|\vx_{t}, \vx_0)$ transition step can act as a ground-truth signal, since it defines how to denoise a noisy image $\vx_t$ with access to what the final, completely denoised image $\vx_0$ should be.  This term is therefore minimized when the two denoising steps match as closely as possible, as measured by their KL Divergence.
\end{enumerate}
As a side note, one observes that in the process of both ELBO derivations (Equation \ref{eq:45} and Equation \ref{eq:51}), only the Markov assumption is used; as a result these formulae will hold true for any arbitrary Markovian HVAE.  Furthermore, when we set $T=1$, both of the ELBO interpretations for a VDM exactly recreate the ELBO equation of a vanilla VAE, as written in Equation \ref{eq:19}.

In this derivation of the ELBO, the bulk of the optimization cost once again lies in the summation term, which dominates the reconstruction term.  Whereas each KL Divergence term $\infdiv{q(\vx_{t-1}|\vx_t, \vx_0)}{p_{\bm{\theta}}(\vx_{t-1}|\vx_t)}$ is difficult to minimize for arbitrary posteriors in arbitrarily complex Markovian HVAEs due to the added complexity of simultaneously learning the encoder, in a VDM we can leverage the Gaussian transition assumption to make optimization tractable.  By Bayes rule, we have:
$$q(\vx_{t-1}|\vx_t, \vx_0) = \frac{q(\vx_t | \vx_{t-1}, \vx_0)q(\vx_{t-1}|\vx_0)}{q(\vx_{t}|\vx_0)}$$
\begin{figure}
  \centering
  \includegraphics[width=\linewidth]{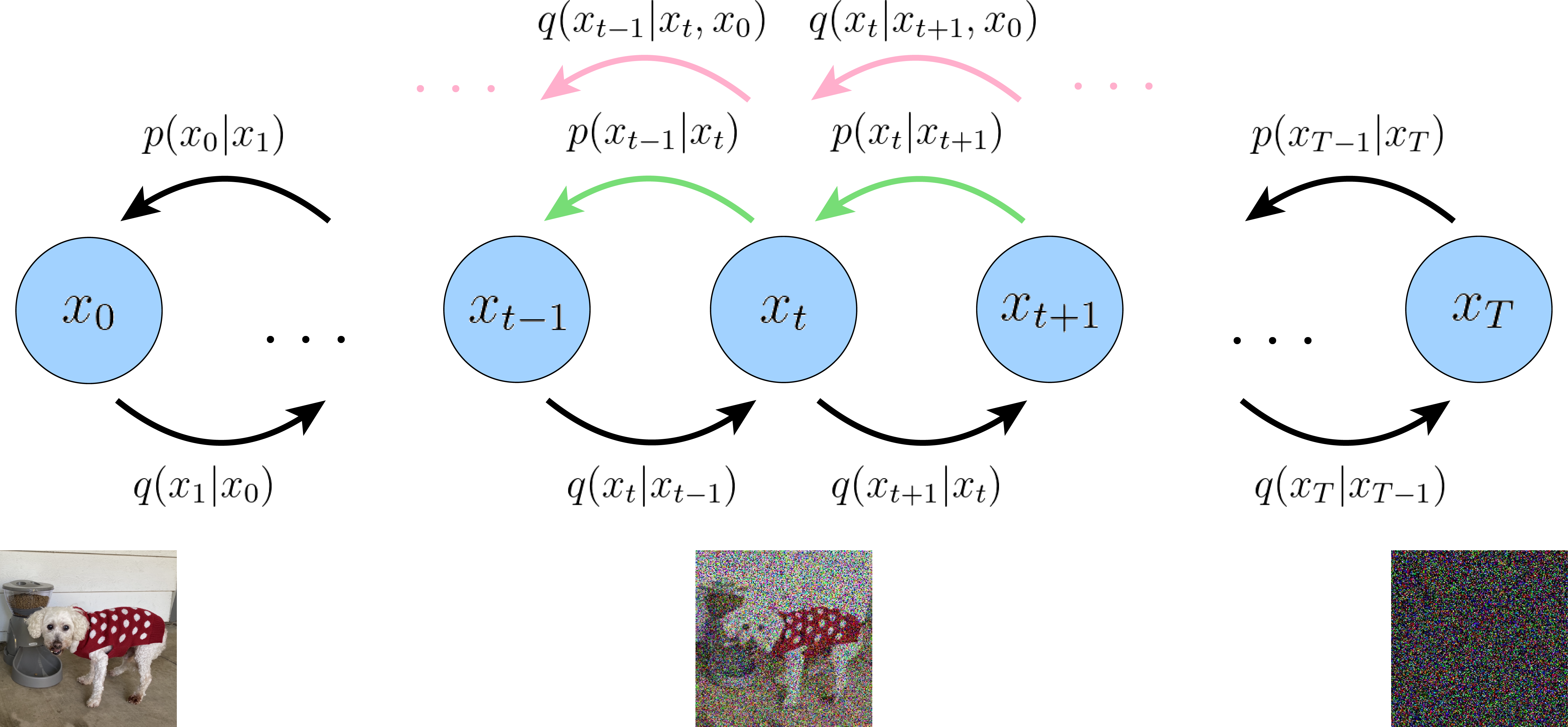}
  \caption{Depicted is an alternate, lower-variance method to optimize a VDM; we compute the form of ground-truth denoising step $q(\vx_{t-1}|\vx_t, \vx_0)$ using Bayes rule, and minimize its KL Divergence with our approximate denoising step $p_{\bm{\theta}}(\vx_{t-1}|\vx_t)$.  This is once again denoted visually by matching the distributions represented by the green arrows with those of the pink arrows.  Artistic liberty is at play here; in the full picture, each pink arrow must also stem from $\vx_0$, as it is also a conditioning term.}
  \label{fig:second_deriv}
\end{figure}As we already know that $q(\vx_t | \vx_{t-1}, \vx_0) = q(\vx_t | \vx_{t-1}) = \mathcal{N}(\vx_{t} ; \sqrt{\alpha_t} \vx_{t-1}, (1 - \alpha_t)\textbf{I})$ from our assumption regarding encoder transitions (Equation \ref{eq:27}), what remains is deriving for the forms of $q(\vx_t|\vx_0)$ and $q(\vx_{t-1}|\vx_0)$.  Fortunately, these are also made tractable by utilizing the fact that the encoder transitions of a VDM are linear Gaussian models.  Recall that under the reparameterization trick, samples $\vx_t \sim q(\vx_t | \vx_{t-1})$ can be rewritten as:
\begin{align}
    \vx_t = \sqrt{\alpha_t}\vx_{t-1} + \sqrt{1 - \alpha_t}\bm{\epsilon} \quad \text{with } \bm{\epsilon} \sim \mathcal{N}(\bm{\epsilon}; \bm{0}, \textbf{I})
\end{align}
and that similarly, samples $\vx_{t-1} \sim q(\vx_{t-1} | \vx_{t-2})$ can be rewritten as:
\begin{align}
    \vx_{t-1} = \sqrt{\alpha_{t-1}}\vx_{t-2} + \sqrt{1 - \alpha_{t-1}}\bm{\epsilon} \quad \text{with } \bm{\epsilon} \sim \mathcal{N}(\bm{\epsilon}; \bm{0}, \textbf{I})
\end{align}
Then, the form of $q(\vx_t|\vx_0)$ can be recursively derived through repeated applications of the reparameterization trick.
Suppose that we have access to 2$T$ random noise variables $\{\bm{\epsilon}_t^*,\bm{\epsilon}_t\}_{t=0}^T \iid \mathcal{N}(\bm{\epsilon}; \bm{0},\textbf{I})$.  Then, for an arbitrary sample $\vx_t \sim q(\vx_t|\vx_0)$, we can rewrite it as:
\begingroup
\begin{align}
\vx_t  &= \sqrt{\alpha_t}\vx_{t-1} + \sqrt{1 - \alpha_t}\bm{\epsilon}_{t-1}^*\\
&= \sqrt{\alpha_t}\left(\sqrt{\alpha_{t-1}}\vx_{t-2} + \sqrt{1 - \alpha_{t-1}}\bm{\epsilon}_{t-2}^*\right) + \sqrt{1 - \alpha_t}\bm{\epsilon}_{t-1}^*\\
&= \sqrt{\alpha_t\alpha_{t-1}}\vx_{t-2} + \sqrt{\alpha_t - \alpha_t\alpha_{t-1}}\bm{\epsilon}_{t-2}^* + \sqrt{1 - \alpha_t}\bm{\epsilon}_{t-1}^*\\
&= \sqrt{\alpha_t\alpha_{t-1}}\vx_{t-2} + \sqrt{\sqrt{\alpha_t - \alpha_t\alpha_{t-1}}^2 + \sqrt{1 - \alpha_t}^2}\bm{\epsilon}_{t-2} \label{eq:63}\\
&= \sqrt{\alpha_t\alpha_{t-1}}\vx_{t-2} + \sqrt{\alpha_t - \alpha_t\alpha_{t-1} + 1 - \alpha_t}\bm{\epsilon}_{t-2}\\
&= \sqrt{\alpha_t\alpha_{t-1}}\vx_{t-2} + \sqrt{1 - \alpha_t\alpha_{t-1}}\bm{\epsilon}_{t-2} \label{eq:66}\\
&= \ldots\\
&= \sqrt{\prod_{i=1}^t\alpha_i}\vx_0 + \sqrt{1 - \prod_{i=1}^t\alpha_i}\bm{\bm{\epsilon}}_0\\
&= \sqrt{\bar\alpha_t}\vx_0 + \sqrt{1 - \bar\alpha_t}\bm{\bm{\epsilon}}_0 \label{eq:68}\\
&\sim \mathcal{N}(\vx_{t} ; \sqrt{\bar\alpha_t}\vx_0, \left(1 - \bar\alpha_t\right)\textbf{I}) \label{eq:61}
\end{align}
where in Equation \ref{eq:63} we have utilized the fact that the \href{https://en.wikipedia.org/wiki/Sum_of_normally_distributed_random_variables}{sum of two independent Gaussian random variables} remains a Gaussian with mean being the sum of the two means, and variance being the sum of the two variances.  Interpreting $\sqrt{1 - \alpha_t}\bm{\epsilon}_{t-1}^*$ as a sample from Gaussian $\mathcal{N}(\bm{0}, (1 - \alpha_t)\textbf{I})$, and $\sqrt{\alpha_t - \alpha_t\alpha_{t-1}}\bm{\epsilon}_{t-2}^*$ as a sample from Gaussian $\mathcal{N}(\bm{0}, (\alpha_t - \alpha_t\alpha_{t-1})\textbf{I})$, we can then treat their sum as a random variable sampled from Gaussian $\mathcal{N}(\bm{0}, (1 - \alpha_t + \alpha_t - \alpha_t\alpha_{t-1})\textbf{I}) = \mathcal{N}(\bm{0}, (1 - \alpha_t\alpha_{t-1})\textbf{I})$.  A sample from this distribution can then be represented using the reparameterization trick as $\sqrt{1 - \alpha_t\alpha_{t-1}}\bm{\epsilon}_{t-2}$, as in Equation \ref{eq:66}.

\newpage
We have therefore derived the Gaussian form of $q(\vx_t|\vx_0)$.  This derivation can be modified to also yield the Gaussian parameterization describing $q(\vx_{t-1}|\vx_0)$.  Now, knowing the forms of both $q(\vx_t|\vx_0)$ and $q(\vx_{t-1}|\vx_0)$, we can proceed to calculate the form of $q(\vx_{t-1}|\vx_t, \vx_0)$ by substituting into the Bayes rule expansion:
\begingroup
\allowdisplaybreaks
\begin{align}
\scalemath{0.94}{q(\vx_{t-1}|\vx_t, \vx_0)}
&= \scalemath{0.94}{\frac{q(\vx_t | \vx_{t-1}, \vx_0)q(\vx_{t-1}|\vx_0)}{q(\vx_{t}|\vx_0)}}\\
&= \scalemath{0.94}{\frac{\mathcal{N}(\vx_{t} ; \sqrt{\alpha_t} \vx_{t-1}, (1 - \alpha_t)\textbf{I})\mathcal{N}(\vx_{t-1} ; \sqrt{\bar\alpha_{t-1}}\vx_0, (1 - \bar\alpha_{t-1}) \textbf{I})}{\mathcal{N}(\vx_{t} ; \sqrt{\bar\alpha_{t}}\vx_0, (1 - \bar\alpha_{t})\textbf{I})}}\\
&\propto \scalemath{0.94}{\text{exp}\left\{-\left[\frac{(\vx_{t} - \sqrt{\alpha_t} \vx_{t-1})^2}{2(1 - \alpha_t)} + \frac{(\vx_{t-1} - \sqrt{\bar\alpha_{t-1}} \vx_0)^2}{2(1 - \bar\alpha_{t-1})} - \frac{(\vx_{t} - \sqrt{\bar\alpha_t} \vx_{0})^2}{2(1 - \bar\alpha_t)} \right]\right\}}\\
&= \scalemath{0.94}{\text{exp}\left\{-\frac{1}{2}\left[\frac{(\vx_{t} - \sqrt{\alpha_t} \vx_{t-1})^2}{1 - \alpha_t} + \frac{(\vx_{t-1} - \sqrt{\bar\alpha_{t-1}} \vx_0)^2}{1 - \bar\alpha_{t-1}} - \frac{(\vx_{t} - \sqrt{\bar\alpha_t} \vx_{0})^2}{1 - \bar\alpha_t} \right]\right\}}\\
&= \scalemath{0.94}{\text{exp}\left\{-\frac{1}{2}\left[\frac{(-2\sqrt{\alpha_t} \vx_{t}\vx_{t-1} + \alpha_t \vx_{t-1}^2)}{1 - \alpha_t} + \frac{(\vx_{t-1}^2 - 2\sqrt{\bar\alpha_{t-1}}\vx_{t-1} \vx_0)}{1 - \bar\alpha_{t-1}} + C(\vx_t, \vx_0)\right]\right\}} \label{eq:73}\\
&\propto \scalemath{0.94}{\text{exp}\left\{-\frac{1}{2}\left[- \frac{2\sqrt{\alpha_t} \vx_{t}\vx_{t-1}}{1 - \alpha_t} + \frac{\alpha_t \vx_{t-1}^2}{1 - \alpha_t} + \frac{\vx_{t-1}^2}{1 - \bar\alpha_{t-1}} - \frac{2\sqrt{\bar\alpha_{t-1}}\vx_{t-1} \vx_0}{1 - \bar\alpha_{t-1}}\right]\right\}}\\
&= \scalemath{0.94}{\text{exp}\left\{-\frac{1}{2}\left[(\frac{\alpha_t}{1 - \alpha_t} + \frac{1}{1 - \bar\alpha_{t-1}})\vx_{t-1}^2 - 2\left(\frac{\sqrt{\alpha_t}\vx_{t}}{1 - \alpha_t} + \frac{\sqrt{\bar\alpha_{t-1}}\vx_0}{1 - \bar\alpha_{t-1}}\right)\vx_{t-1}\right]\right\}}\\
&= \scalemath{0.94}{\text{exp}\left\{-\frac{1}{2}\left[\frac{\alpha_t(1-\bar\alpha_{t-1}) + 1 - \alpha_t}{(1 - \alpha_t)(1 - \bar\alpha_{t-1})}\vx_{t-1}^2 - 2\left(\frac{\sqrt{\alpha_t}\vx_{t}}{1 - \alpha_t} + \frac{\sqrt{\bar\alpha_{t-1}}\vx_0}{1 - \bar\alpha_{t-1}}\right)\vx_{t-1}\right]\right\}}\\
&= \scalemath{0.94}{\text{exp}\left\{-\frac{1}{2}\left[\frac{\alpha_t-\bar\alpha_{t} + 1 - \alpha_t}{(1 - \alpha_t)(1 - \bar\alpha_{t-1})}\vx_{t-1}^2 - 2\left(\frac{\sqrt{\alpha_t}\vx_{t}}{1 - \alpha_t} + \frac{\sqrt{\bar\alpha_{t-1}}\vx_0}{1 - \bar\alpha_{t-1}}\right)\vx_{t-1}\right]\right\}}\\
&= \scalemath{0.94}{\text{exp}\left\{-\frac{1}{2}\left[\frac{1 -\bar\alpha_{t}}{(1 - \alpha_t)(1 - \bar\alpha_{t-1})}\vx_{t-1}^2 - 2\left(\frac{\sqrt{\alpha_t}\vx_{t}}{1 - \alpha_t} + \frac{\sqrt{\bar\alpha_{t-1}}\vx_0}{1 - \bar\alpha_{t-1}}\right)\vx_{t-1}\right]\right\}}\\
&= \scalemath{0.94}{\text{exp}\left\{-\frac{1}{2}\left(\frac{1 -\bar\alpha_{t}}{(1 - \alpha_t)(1 - \bar\alpha_{t-1})}\right)\left[\vx_{t-1}^2 - 2\frac{\left(\frac{\sqrt{\alpha_t}\vx_{t}}{1 - \alpha_t} + \frac{\sqrt{\bar\alpha_{t-1}}\vx_0}{1 - \bar\alpha_{t-1}}\right)}{\frac{1 -\bar\alpha_{t}}{(1 - \alpha_t)(1 - \bar\alpha_{t-1})}}\vx_{t-1}\right]\right\}}\\
&= \scalemath{0.94}{\text{exp}\left\{-\frac{1}{2}\left(\frac{1 -\bar\alpha_{t}}{(1 - \alpha_t)(1 - \bar\alpha_{t-1})}\right)\left[\vx_{t-1}^2 - 2\frac{\left(\frac{\sqrt{\alpha_t}\vx_{t}}{1 - \alpha_t} + \frac{\sqrt{\bar\alpha_{t-1}}\vx_0}{1 - \bar\alpha_{t-1}}\right)(1 - \alpha_t)(1 - \bar\alpha_{t-1})}{1 -\bar\alpha_{t}}\vx_{t-1}\right]\right\}}\\
&= \scalemath{0.94}{\text{exp}\left\{-\frac{1}{2}\left(\frac{1}{\frac{(1 - \alpha_t)(1 - \bar\alpha_{t-1})}{1 -\bar\alpha_{t}}}\right)\left[\vx_{t-1}^2 - 2\frac{\sqrt{\alpha_t}(1-\bar\alpha_{t-1})\vx_{t} + \sqrt{\bar\alpha_{t-1}}(1-\alpha_t)\vx_0}{1 -\bar\alpha_{t}}\vx_{t-1}\right]\right\}}\\
&\propto \scalemath{0.94}{\mathcal{N}(\vx_{t-1} ;} \underbrace{\scalemath{0.94}{\frac{\sqrt{\alpha_t}(1-\bar\alpha_{t-1})\vx_{t} + \sqrt{\bar\alpha_{t-1}}(1-\alpha_t)\vx_0}{1 -\bar\alpha_{t}}}}_{\mu_q(\vx_t, \vx_0)}, \underbrace{\scalemath{0.94}{\frac{(1 - \alpha_t)(1 - \bar\alpha_{t-1})}{1 -\bar\alpha_{t}}\textbf{I}}}_{\bm{\Sigma}_q(t)}) \label{eq:78}
\end{align}
\endgroup
where in Equation \ref{eq:73}, $C(\vx_t, \vx_0)$ is a constant term with respect to $\vx_{t-1}$ computed as a combination of only $\vx_t$, $\vx_0$, and $\alpha$ values; this term is implicitly returned in Equation \ref{eq:78} to complete the square.

We have therefore shown that at each step, $\vx_{t-1} \sim q(\vx_{t-1}| \vx_t, \vx_0)$ is normally distributed, with mean $\bm{\mu}_q(\vx_t, \vx_0)$ that is a function of $\vx_t$ and $\vx_0$, and variance $\bm{\Sigma}_q(t)$ as a function of $\alpha$ coefficients.  These $\alpha$ coefficients are known and fixed at each timestep; they are either set permanently when modeled as hyperparameters, or treated as the current inference output of a network that seeks to model them.  Following Equation \ref{eq:78}, we can rewrite our variance equation as $\bm{\Sigma}_q(t) = \sigma_q^2(t)\textbf{I}$, where:
\begin{align}
    \sigma_q^2(t) = \frac{(1 - \alpha_t)(1 - \bar\alpha_{t-1})}{1 -\bar\alpha_{t}} \label{eq:79}
\end{align}
In order to match approximate denoising transition step $p_{\bm{\theta}}(\vx_{t-1}|\vx_t)$ to ground-truth denoising transition step $q(\vx_{t-1}| \vx_t, \vx_0)$ as closely as possible, we can also model it as a Gaussian.  Furthermore, as all $\alpha$ terms are known to be frozen at each timestep, we can immediately construct the variance of the approximate denoising transition step to also be $\bm{\Sigma}_q(t) = \sigma_q^2(t)\textbf{I}$.  We must parameterize its mean $\bm{\mu}_{\bm{\theta}}(\vx_t, t)$ as a function of $\vx_t$, however, since $p_{\bm{\theta}}(\vx_{t-1}|\vx_t)$ does not condition on $\vx_0$.

Recall that the \href{https://en.wikipedia.org/wiki/Kullback\%E2\%80\%93Leibler_divergence#Multivariate_normal_distributions}{KL Divergence between two Gaussian distributions} is:
\begin{align}
\infdiv{\mathcal{N}(\vx; \bm{\mu}_x,\bm{\Sigma}_x)}{\mathcal{N}(\vy; \bm{\mu}_y,\bm{\Sigma}_y)}
&=\frac{1}{2}\left[\log\frac{|\bm{\Sigma}_y|}{|\bm{\Sigma}_x|} - d + \text{tr}(\bm{\Sigma}_y^{-1}\bm{\Sigma}_x)
+ (\bm{\mu}_y-\bm{\mu}_x)^T \bm{\Sigma}_y^{-1} (\bm{\mu}_y-\bm{\mu}_x)\right]
\end{align}
In our case, where we can set the variances of the two Gaussians to match exactly, optimizing the KL Divergence term reduces to minimizing the difference between the means of the two distributions:
\begin{align}
& \quad \,\argmin_{\bm{\theta}} \infdiv{q(\vx_{t-1}|\vx_t, \vx_0)}{p_{\bm{\theta}}(\vx_{t-1}|\vx_t)} \nonumber \\
&= \argmin_{\bm{\theta}}\infdiv{\mathcal{N}(\vx_{t-1}; \bm{\mu}_q,\bm{\Sigma}_q(t))}{\mathcal{N}(\vx_{t-1}; \bm{\mu}_{\bm{\theta}},\bm{\Sigma}_q(t))}\\
&=\argmin_{\bm{\theta}}\frac{1}{2}\left[\log\frac{|\bm{\Sigma}_q(t)|}{|\bm{\Sigma}_q(t)|} - d + \text{tr}(\bm{\Sigma}_q(t)^{-1}\bm{\Sigma}_q(t))
+ (\bm{\mu}_{\bm{\theta}}-\bm{\mu}_q)^T \bm{\Sigma}_q(t)^{-1} (\bm{\mu}_{\bm{\theta}}-\bm{\mu}_q)\right]\\
&=\argmin_{\bm{\theta}}\frac{1}{2}\left[\log1 - d + d + (\bm{\mu}_{\bm{\theta}}-\bm{\mu}_q)^T \bm{\Sigma}_q(t)^{-1} (\bm{\mu}_{\bm{\theta}}-\bm{\mu}_q)\right]\\
&=\argmin_{\bm{\theta}}\frac{1}{2}\left[(\bm{\mu}_{\bm{\theta}}-\bm{\mu}_q)^T \bm{\Sigma}_q(t)^{-1} (\bm{\mu}_{\bm{\theta}}-\bm{\mu}_q)\right]\\
&=\argmin_{\bm{\theta}}\frac{1}{2}\left[(\bm{\mu}_{\bm{\theta}}-\bm{\mu}_q)^T \left(\sigma_q^2(t)\textbf{I}\right)^{-1} (\bm{\mu}_{\bm{\theta}}-\bm{\mu}_q)\right]\\
&=\argmin_{\bm{\theta}}\frac{1}{2\sigma_q^2(t)}\left[\left\lVert\bm{\mu}_{\bm{\theta}}-\bm{\mu}_q\right\rVert_2^2\right]
\end{align}
where we have written $\bm{\mu}_q$ as shorthand for $\bm{\mu}_q(\vx_t, \vx_0)$, and $\bm{\mu}_{\bm{\theta}}$ as shorthand for $\bm{\mu}_{\bm{\theta}}(\vx_t, t)$ for brevity.  In other words, we want to optimize a $\bm{\mu}_{\bm{\theta}}(\vx_t, t)$ that matches $\bm{\mu}_q(\vx_t, \vx_0)$, which from our derived Equation \ref{eq:78}, takes the form: 
\begin{align}
    \bm{\mu}_q(\vx_t, \vx_0) = \frac{\sqrt{\alpha_t}(1-\bar\alpha_{t-1})\vx_{t} + \sqrt{\bar\alpha_{t-1}}(1-\alpha_t)\vx_0}{1 -\bar\alpha_{t}}
\end{align}
As $\bm{\mu}_{\bm{\theta}}(\vx_t, t)$ also conditions on $\vx_t$, we can match $\bm{\mu}_q(\vx_t, \vx_0)$ closely by setting it to the following form:
\begin{align}
    \bm{\mu}_{\bm{\theta}}(\vx_t, t) = \frac{\sqrt{\alpha_t}(1-\bar\alpha_{t-1})\vx_{t} + \sqrt{\bar\alpha_{t-1}}(1-\alpha_t)\hat \vx_{\bm{\theta}}(\vx_t, t)}{1 -\bar\alpha_{t}}
\end{align}
where $\hat \vx_{\bm{\theta}}(\vx_t, t)$ is parameterized by a neural network that seeks to predict $\vx_0$ from noisy image $\vx_t$ and time index $t$. Then, the optimization problem simplifies to:
\begingroup
\allowdisplaybreaks
\begin{align}
& \scalemath{0.93}{\quad \,\argmin_{\bm{\theta}} \infdiv{q(\vx_{t-1}|\vx_t, \vx_0)}{p_{\bm{\theta}}(\vx_{t-1}|\vx_t)}} \nonumber \\
&= \scalemath{0.93}{\argmin_{\bm{\theta}}\infdiv{\mathcal{N}\left(\vx_{t-1}; \bm{\mu}_q,\bm{\Sigma}_q\left(t\right)\right)}{\mathcal{N}\left(\vx_{t-1}; \bm{\mu}_{\bm{\theta}},\bm{\Sigma}_q\left(t\right)\right)}}\\
&=\scalemath{0.93}{\argmin_{\bm{\theta}}\frac{1}{2\sigma_q^2(t)}\left[\left\lVert\frac{\sqrt{\alpha_t}(1-\bar\alpha_{t-1})\vx_{t} + \sqrt{\bar\alpha_{t-1}}(1-\alpha_t)\hat \vx_{\bm{\theta}}(\vx_t, t)}{1 -\bar\alpha_{t}}-\frac{\sqrt{\alpha_t}(1-\bar\alpha_{t-1})\vx_{t} + \sqrt{\bar\alpha_{t-1}}(1-\alpha_t)\vx_0}{1 -\bar\alpha_{t}}\right\rVert_2^2\right]}\\
&=\scalemath{0.93}{\argmin_{\bm{\theta}}\frac{1}{2\sigma_q^2(t)}\left[\left\lVert\frac{\sqrt{\bar\alpha_{t-1}}(1-\alpha_t)\hat \vx_{\bm{\theta}}(\vx_t, t)}{1 -\bar\alpha_{t}}-\frac{\sqrt{\bar\alpha_{t-1}}(1-\alpha_t)\vx_0}{1 -\bar\alpha_{t}}\right\rVert_2^2\right]}\\
&=\scalemath{0.93}{\argmin_{\bm{\theta}}\frac{1}{2\sigma_q^2(t)}\left[\left\lVert\frac{\sqrt{\bar\alpha_{t-1}}(1-\alpha_t)}{1 -\bar\alpha_{t}}\left(\hat \vx_{\bm{\theta}}(\vx_t, t) - \vx_0\right)\right\rVert_2^2\right]}\\
&=\scalemath{0.93}{\argmin_{\bm{\theta}}\frac{1}{2\sigma_q^2(t)}\frac{\bar\alpha_{t-1}(1-\alpha_t)^2}{(1 -\bar\alpha_{t})^2}\left[\left\lVert\hat \vx_{\bm{\theta}}(\vx_t, t) - \vx_0\right\rVert_2^2\right]} \label{eq:93}
\end{align}
\endgroup
Therefore, optimizing a VDM boils down to learning a neural network to predict the original ground truth image from an arbitrarily noisified version of it~\cite{ho2020denoising}.  Furthermore, minimizing the summation term of our derived ELBO objective (Equation \ref{eq:51}) across all noise levels can be approximated by minimizing the expectation over all timesteps:
\begin{align}
\argmin_{\bm{\theta}}\mathbb{E}_{t\sim U\{2, T\}}\left[\mathbb{E}_{q(\vx_{t}|\vx_0)}\left[\infdiv{q(\vx_{t-1}|\vx_t, \vx_0)}{p_{\bm{\theta}}(\vx_{t-1}|\vx_t)}\right]\right] \label{eq:94}
\end{align}
which can then be optimized using stochastic samples over timesteps.

\subsubsection*{Learning Diffusion Noise Parameters}
\addcontentsline{toc}{section}{\protect\numberline{}\protect\numberline{}Learning Diffusion Noise Parameters}%
Let us investigate how the noise parameters of a VDM can be jointly learned.  One potential approach is to model $\alpha_t$ using a neural network $\hat\alpha_{\bm{\eta}}(t)$ with parameters $\bm{\eta}$.  However, this is inefficient as inference must be performed multiple times at each timestep $t$ to compute $\bar\alpha_t$.  Whereas caching can mitigate this computational cost, we can also derive an alternate way to learn the diffusion noise parameters. By substituting our variance equation from Equation \ref{eq:79} into our derived per-timestep objective in Equation \ref{eq:93}, we can reduce:
\begin{align}
\scalemath{0.9}{\frac{1}{2\sigma_q^2(t)}\frac{\bar\alpha_{t-1}(1-\alpha_t)^2}{(1 -\bar\alpha_{t})^2}\left[\left\lVert\hat \vx_{\bm{\theta}}(\vx_t, t) - \vx_0\right\rVert_2^2\right]}
&= \scalemath{0.9}{\frac{1}{2\frac{(1 - \alpha_t)(1 - \bar\alpha_{t-1})}{1 -\bar\alpha_{t}}}\frac{\bar\alpha_{t-1}(1-\alpha_t)^2}{(1 -\bar\alpha_{t})^2}\left[\left\lVert\hat \vx_{\bm{\theta}}(\vx_t, t) - \vx_0\right\rVert_2^2\right]}\\
&= \scalemath{0.9}{\frac{1}{2}\frac{1 -\bar\alpha_{t}}{(1 - \alpha_t)(1 - \bar\alpha_{t-1})}\frac{\bar\alpha_{t-1}(1-\alpha_t)^2}{(1 -\bar\alpha_{t})^2}\left[\left\lVert\hat \vx_{\bm{\theta}}(\vx_t, t) - \vx_0\right\rVert_2^2\right]}\\
&= \scalemath{0.9}{\frac{1}{2}\frac{\bar\alpha_{t-1}(1-\alpha_t)}{(1 - \bar\alpha_{t-1})(1 -\bar\alpha_{t})}\left[\left\lVert\hat \vx_{\bm{\theta}}(\vx_t, t) - \vx_0\right\rVert_2^2\right]}\\
&= \scalemath{0.9}{\frac{1}{2}\frac{\bar\alpha_{t-1}-\bar\alpha_t}{(1 - \bar\alpha_{t-1})(1 -\bar\alpha_{t})}\left[\left\lVert\hat \vx_{\bm{\theta}}(\vx_t, t) - \vx_0\right\rVert_2^2\right]}\\
&= \scalemath{0.9}{\frac{1}{2}\frac{\bar\alpha_{t-1} - \bar\alpha_{t-1}\bar\alpha_t + \bar\alpha_{t-1}\bar\alpha_t-\bar\alpha_t}{(1 - \bar\alpha_{t-1})(1 -\bar\alpha_{t})}\left[\left\lVert\hat \vx_{\bm{\theta}}(\vx_t, t) - \vx_0\right\rVert_2^2\right]}\\
&= \scalemath{0.9}{\frac{1}{2}\frac{\bar\alpha_{t-1}(1 - \bar\alpha_t) -\bar\alpha_t(1 - \bar\alpha_{t-1})}{(1 - \bar\alpha_{t-1})(1 -\bar\alpha_{t})}\left[\left\lVert\hat \vx_{\bm{\theta}}(\vx_t, t) - \vx_0\right\rVert_2^2\right]}\\
&= \scalemath{0.9}{\frac{1}{2}\left(\frac{\bar\alpha_{t-1}(1 - \bar\alpha_t)}{(1 - \bar\alpha_{t-1})(1 -\bar\alpha_{t})} -\frac{\bar\alpha_t(1 - \bar\alpha_{t-1})}{(1 - \bar\alpha_{t-1})(1 -\bar\alpha_{t})}\right)\left[\left\lVert\hat \vx_{\bm{\theta}}(\vx_t, t) - \vx_0\right\rVert_2^2\right]}\\
&= \scalemath{0.9}{\frac{1}{2}\left(\frac{\bar\alpha_{t-1}}{1 - \bar\alpha_{t-1}} -\frac{\bar\alpha_t}{1 -\bar\alpha_{t}}\right)\left[\left\lVert\hat \vx_{\bm{\theta}}(\vx_t, t) - \vx_0\right\rVert_2^2\right]} \label{eq:102}
\end{align}
Recall from Equation \ref{eq:61} that $q(\vx_t|\vx_0)$ is a Gaussian of form $\mathcal{N}(\vx_{t} ; \sqrt{\bar\alpha_t}\vx_0, \left(1 - \bar\alpha_t\right)\textbf{I})$.  Then, following the definition of the \href{https://en.wikipedia.org/wiki/Signal-to-noise_ratio#Alternate_definition}{signal-to-noise ratio (SNR)} as $\text{SNR} = \frac{\mu^2}{\sigma^2}$, we can write the SNR at each timestep $t$ as:
\begin{align}
    \text{SNR}(t) &= \frac{\bar\alpha_t}{1 -\bar\alpha_{t}} \label{eq:108}
\end{align}
Then, our derived Equation \ref{eq:102} (and Equation \ref{eq:93}) can be simplified as:
\begin{align}
\frac{1}{2\sigma_q^2(t)}\frac{\bar\alpha_{t-1}(1-\alpha_t)^2}{(1 -\bar\alpha_{t})^2}\left[\left\lVert\hat \vx_{\bm{\theta}}(\vx_t, t) - \vx_0\right\rVert_2^2\right] &= \frac{1}{2}\left(\text{SNR}(t-1) -\text{SNR}(t)\right)\left[\left\lVert\hat \vx_{\bm{\theta}}(\vx_t, t) - \vx_0\right\rVert_2^2\right] \label{eq:104}
\end{align}
As the name implies, the SNR represents the ratio between the original signal and the amount of noise present; a higher SNR represents more signal and a lower SNR represents more noise.  In a diffusion model, we require the SNR to monotonically decrease as timestep $t$ increases; this formalizes the notion that perturbed input $\vx_t$ becomes increasingly noisy over time, until it becomes identical to a standard Gaussian at $t=T$.

Following the simplification of the objective in Equation \ref{eq:104}, we can directly parameterize the SNR at each timestep using a neural network, and learn it jointly along with the diffusion model.  As the SNR must monotonically decrease over time, we can represent it as:
\begin{align}
    \text{SNR}(t) = \text{exp}(-\omega_{\bm{\eta}}(t)) \label{eq:105}
\end{align}
where $\omega_{\bm{\eta}}(t)$ is modeled as a monotonically increasing neural network with parameters $\bm{\eta}$.  Negating $\omega_{\bm{\eta}}(t)$ results in a monotonically decreasing function, whereas the exponential forces the resulting term to be positive.
Note that the objective in Equation \ref{eq:94} must now optimize over $\bm{\eta}$ as well.  By combining our parameterization of SNR in Equation \ref{eq:105} with our definition of SNR in Equation \ref{eq:108}, we can also explicitly derive elegant forms for the value of $\bar\alpha_t$ as well as for the value of $1 - \bar\alpha_t$:
\begin{align}
    &\frac{\bar\alpha_t}{1 -\bar\alpha_{t}} = \text{exp}(-\omega_{\bm{\eta}}(t))\\
    &\therefore \bar\alpha_t = \text{sigmoid}(-\omega_{\bm{\eta}}(t))\\
    &\therefore 1 - \bar\alpha_t = \text{sigmoid}(\omega_{\bm{\eta}}(t))
\end{align}
These terms are necessary for a variety of computations; for example, during optimization, they are used to create arbitrarily noisy $\vx_t$ from input $\vx_0$ using the reparameterization trick, as derived in Equation \ref{eq:68}.
\subsubsection*{Three Equivalent Interpretations}
\addcontentsline{toc}{section}{\protect\numberline{}\protect\numberline{}Three Equivalent Interpretations}%
As we previously proved, a Variational Diffusion Model can be trained by simply learning a neural network to predict the original natural image $\vx_0$ from an arbitrary noised version $\vx_t$ and its time index $t$.  However, $\vx_0$ has two other equivalent parameterizations, which leads to two further interpretations for a VDM.

Firstly, we can utilize the reparameterization trick.  In our derivation of the form of $q(\vx_t|\vx_0)$, we can rearrange Equation \ref{eq:68} to show that:
\begin{align}
\vx_0 &= \frac{\vx_t - \sqrt{1 - \bar\alpha_t}\bm{\epsilon}_0}{\sqrt{\bar\alpha_t}} \label{eq:62}
\end{align}
Plugging this into our previously derived true denoising transition mean $\bm{\mu}_q(\vx_t, \vx_0)$, we can rederive as:
\begingroup
\allowdisplaybreaks
\begin{align}
\bm{\mu}_q(\vx_t, \vx_0) &= \frac{\sqrt{\alpha_t}(1-\bar\alpha_{t-1})\vx_{t} + \sqrt{\bar\alpha_{t-1}}(1-\alpha_t)\vx_0}{1 -\bar\alpha_{t}}\\
&= \frac{\sqrt{\alpha_t}(1-\bar\alpha_{t-1})\vx_{t} + \sqrt{\bar\alpha_{t-1}}(1-\alpha_t)\frac{\vx_t - \sqrt{1 - \bar\alpha_t}\bm{\epsilon}_0}{\sqrt{\bar\alpha_t}}}{1 -\bar\alpha_{t}}\\
&= \frac{\sqrt{\alpha_t}(1-\bar\alpha_{t-1})\vx_{t} + (1-\alpha_t)\frac{\vx_t - \sqrt{1 - \bar\alpha_t}\bm{\epsilon}_0}{\sqrt{\alpha_t}}}{1 -\bar\alpha_{t}}\\
&= \frac{\sqrt{\alpha_t}(1-\bar\alpha_{t-1})\vx_{t}}{1 - \bar\alpha_t} + \frac{(1-\alpha_t)\vx_t}{(1-\bar\alpha_t)\sqrt{\alpha_t}} - \frac{(1 - \alpha_t)\sqrt{1 - \bar\alpha_t}\bm{\epsilon}_0}{(1-\bar\alpha_t)\sqrt{\alpha_t}}\\
&= \left(\frac{\sqrt{\alpha_t}(1-\bar\alpha_{t-1})}{1 - \bar\alpha_t} + \frac{1-\alpha_t}{(1-\bar\alpha_t)\sqrt{\alpha_t}}\right)\vx_t - \frac{(1 - \alpha_t)\sqrt{1 - \bar\alpha_t}}{(1-\bar\alpha_t)\sqrt{\alpha_t}}\bm{\epsilon}_0\\
&= \left(\frac{\alpha_t(1-\bar\alpha_{t-1})}{(1 - \bar\alpha_t)\sqrt{\alpha_t}} + \frac{1-\alpha_t}{(1-\bar\alpha_t)\sqrt{\alpha_t}}\right)\vx_t - \frac{1 - \alpha_t}{\sqrt{1 - \bar\alpha_t}\sqrt{\alpha_t}}\bm{\epsilon}_0\\
&= \frac{\alpha_t-\bar\alpha_{t} + 1-\alpha_t}{(1 - \bar\alpha_t)\sqrt{\alpha_t}}\vx_t - \frac{1 - \alpha_t}{\sqrt{1 - \bar\alpha_t}\sqrt{\alpha_t}}\bm{\epsilon}_0\\
&= \frac{1-\bar\alpha_t}{(1 - \bar\alpha_t)\sqrt{\alpha_t}}\vx_t - \frac{1 - \alpha_t}{\sqrt{1 - \bar\alpha_t}\sqrt{\alpha_t}}\bm{\epsilon}_0\\
&= \frac{1}{\sqrt{\alpha_t}}\vx_t - \frac{1 - \alpha_t}{\sqrt{1 - \bar\alpha_t}\sqrt{\alpha_t}}\bm{\epsilon}_0
\end{align}
Therefore, we can set our approximate denoising transition mean $\bm{\mu}_{\bm{\theta}}(\vx_t, t)$ as:
\begin{align}
\bm{\mu}_{\bm{\theta}}(\vx_t, t) &= \frac{1}{\sqrt{\alpha_t}}\vx_t - \frac{1 - \alpha_t}{\sqrt{1 - \bar\alpha_t}\sqrt{\alpha_t}}\bm{\hat\epsilon}_{\bm{\theta}}(\vx_t, t)
\end{align}
and the corresponding optimization problem becomes:
\begin{align}
& \quad \,\argmin_{\bm{\theta}} \infdiv{q(\vx_{t-1}|\vx_t, \vx_0)}{p_{\bm{\theta}}(\vx_{t-1}|\vx_t)} \nonumber \\
&= \argmin_{\bm{\theta}}\infdiv{\mathcal{N}\left(\vx_{t-1}; \bm{\mu}_q,\bm{\Sigma}_q\left(t\right)\right)}{\mathcal{N}\left(\vx_{t-1}; \bm{\mu}_{\bm{\theta}},\bm{\Sigma}_q\left(t\right)\right)}\\
&=\argmin_{\bm{\theta}}\frac{1}{2\sigma_q^2(t)}\left[\left\lVert\frac{1}{\sqrt{\alpha_t}}\vx_t - \frac{1 - \alpha_t}{\sqrt{1 - \bar\alpha_t}\sqrt{\alpha_t}}\bm{\hat\epsilon}_{\bm{\theta}}(\vx_t, t) - 
\frac{1}{\sqrt{\alpha_t}}\vx_t + \frac{1 - \alpha_t}{\sqrt{1 - \bar\alpha_t}\sqrt{\alpha_t}}\bm{\epsilon}_0\right\rVert_2^2\right]\\
&=\argmin_{\bm{\theta}}\frac{1}{2\sigma_q^2(t)}\left[\left\lVert \frac{1 - \alpha_t}{\sqrt{1 - \bar\alpha_t}\sqrt{\alpha_t}}\bm{\epsilon}_0 - \frac{1 - \alpha_t}{\sqrt{1 - \bar\alpha_t}\sqrt{\alpha_t}}\bm{\hat\epsilon}_{\bm{\theta}}(\vx_t, t)\right\rVert_2^2\right]\\
&=\argmin_{\bm{\theta}}\frac{1}{2\sigma_q^2(t)}\left[\left\lVert \frac{1 - \alpha_t}{\sqrt{1 - \bar\alpha_t}\sqrt{\alpha_t}}(\bm{\epsilon}_0 - \bm{\hat\epsilon}_{\bm{\theta}}(\vx_t, t))\right\rVert_2^2\right]\\
&=\argmin_{\bm{\theta}}\frac{1}{2\sigma_q^2(t)}\frac{(1 - \alpha_t)^2}{(1 - \bar\alpha_t)\alpha_t}\left[\left\lVert\bm{\epsilon}_0 - \bm{\hat\epsilon}_{\bm{\theta}}(\vx_t, t)\right\rVert_2^2\right]
\end{align}
\endgroup
Here, $\bm{\hat\epsilon}_{\bm{\theta}}(\vx_t, t)$ is a neural network that learns to predict the source noise $\bm{\epsilon}_0 \sim \mathcal{N}(\bm{\epsilon}; \bm{0}, \textbf{I})$ that determines $\vx_t$ from $\vx_0$.  We have therefore shown that learning a VDM by predicting the original image $\vx_0$ is equivalent to learning to predict the noise; empirically, however, some works have found that predicting the noise resulted in better performance \cite{ho2020denoising, saharia2022photorealistic}.

To derive the third common interpretation of Variational Diffusion Models, we appeal to Tweedie's Formula~\cite{efron2011tweedie}.  In English, Tweedie's Formula states that the true mean of an exponential family distribution, given samples drawn from it, can be estimated by the maximum likelihood estimate of the samples (aka empirical mean) plus some correction term involving the score of the estimate.  In the case of just one observed sample, the empirical mean is just the sample itself.  It is commonly used to mitigate sample bias; if observed samples all lie on one end of the underlying distribution, then the negative score becomes large and corrects the naive maximum likelihood estimate of the samples towards the true mean.

Mathematically, for a Gaussian variable $\vz \sim \mathcal{N}(\vz;\bm{\mu}_z, \bm{\Sigma}_z)$, Tweedie's Formula states that: 
$$\mathbb{E}\left[\bm{\mu}_z|\vz\right] = \vz + \bm{\Sigma}_z\nabla_\vz\log p(\vz)$$
In this case, we apply it to predict the true posterior mean of $\vx_t$ given its samples.  From Equation \ref{eq:61}, we know that:
$$q(\vx_t|\vx_0) = \mathcal{N}(\vx_{t} ; \sqrt{\bar\alpha_t}\vx_0, \left(1 - \bar\alpha_t\right)\textbf{I})$$
Then, by Tweedie's Formula, we have:
\begin{align}
\mathbb{E}\left[\bm{\mu}_{x_t}|\vx_t\right] = \vx_t + (1 - \bar\alpha_t)\nabla_{\vx_t}\log p(\vx_t)
\end{align}
where we write $\nabla_{\vx_t}\log p(\vx_t)$ as $\nabla\log p(\vx_t)$ for notational simplicity.
According to Tweedie’s Formula, the best estimate for the true mean that $\vx_t$ is generated from, $\bm{\mu}_{x_t} = \sqrt{\bar\alpha_t}\vx_0$, is defined as:
\begin{align}
    \sqrt{\bar\alpha_t}\vx_0 = \vx_t + (1 - \bar\alpha_t)\nabla\log p(\vx_t)\\
    \therefore \vx_0 = \frac{\vx_t + (1 - \bar\alpha_t)\nabla\log p(\vx_t)}{\sqrt{\bar\alpha_t}} \label{eq:109}
\end{align}
Then, we can plug Equation \ref{eq:109} into our ground-truth denoising transition mean $\bm{\mu}_q(\vx_t, \vx_0)$ once again and derive a new form:
\begin{align}
\bm{\mu}_q(\vx_t, \vx_0) &= \frac{\sqrt{\alpha_t}(1-\bar\alpha_{t-1})\vx_{t} + \sqrt{\bar\alpha_{t-1}}(1-\alpha_t)\vx_0}{1 -\bar\alpha_{t}}\\
&= \frac{\sqrt{\alpha_t}(1-\bar\alpha_{t-1})\vx_{t} + \sqrt{\bar\alpha_{t-1}}(1-\alpha_t)\frac{\vx_t + (1 - \bar\alpha_t)\nabla\log p(\vx_t)}{\sqrt{\bar\alpha_t}}}{1 -\bar\alpha_{t}}\\
&= \frac{\sqrt{\alpha_t}(1-\bar\alpha_{t-1})\vx_{t} + (1-\alpha_t)\frac{\vx_t + (1 - \bar\alpha_t)\nabla\log p(\vx_t)}{\sqrt{\alpha_t}}}{1 -\bar\alpha_{t}}\\
&= \frac{\sqrt{\alpha_t}(1-\bar\alpha_{t-1})\vx_{t}}{1 - \bar\alpha_t} + \frac{(1-\alpha_t)\vx_t}{(1-\bar\alpha_t)\sqrt{\alpha_t}} + \frac{(1 - \alpha_t)(1 - \bar\alpha_t)\nabla\log p(\vx_t)}{(1-\bar\alpha_t)\sqrt{\alpha_t}}\\
&= \left(\frac{\sqrt{\alpha_t}(1-\bar\alpha_{t-1})}{1 - \bar\alpha_t} + \frac{1-\alpha_t}{(1-\bar\alpha_t)\sqrt{\alpha_t}}\right)\vx_t + \frac{1 - \alpha_t}{\sqrt{\alpha_t}}\nabla\log p(\vx_t)\\
&= \left(\frac{\alpha_t(1-\bar\alpha_{t-1})}{(1 - \bar\alpha_t)\sqrt{\alpha_t}} + \frac{1-\alpha_t}{(1-\bar\alpha_t)\sqrt{\alpha_t}}\right)\vx_t + \frac{1 - \alpha_t}{\sqrt{\alpha_t}}\nabla\log p(\vx_t)\\
&= \frac{\alpha_t-\bar\alpha_{t} + 1-\alpha_t}{(1 - \bar\alpha_t)\sqrt{\alpha_t}}\vx_t + \frac{1 - \alpha_t}{\sqrt{\alpha_t}}\nabla\log p(\vx_t)\\
&= \frac{1-\bar\alpha_t}{(1 - \bar\alpha_t)\sqrt{\alpha_t}}\vx_t + \frac{1 - \alpha_t}{\sqrt{\alpha_t}}\nabla\log p(\vx_t)\\
&= \frac{1}{\sqrt{\alpha_t}}\vx_t + \frac{1 - \alpha_t}{\sqrt{\alpha_t}}\nabla\log p(\vx_t)
\end{align}
Therefore, we can also set our approximate denoising transition mean $\bm{\mu}_{\bm{\theta}}(\vx_t, t)$ as:
\begin{align}
\bm{\mu}_{\bm{\theta}}(\vx_t, t) &= \frac{1}{\sqrt{\alpha_t}}\vx_t + \frac{1 - \alpha_t}{\sqrt{\alpha_t}}\vs_{\bm{\theta}}(\vx_t, t)
\end{align}
and the corresponding optimization problem becomes:
\begin{align}
& \quad \,\argmin_{\bm{\theta}} \infdiv{q(\vx_{t-1}|\vx_t, \vx_0)}{p_{\bm{\theta}}(\vx_{t-1}|\vx_t)} \nonumber \\
&= \argmin_{\bm{\theta}}\infdiv{\mathcal{N}\left(\vx_{t-1}; \bm{\mu}_q,\bm{\Sigma}_q\left(t\right)\right)}{\mathcal{N}\left(\vx_{t-1}; \bm{\mu}_{\bm{\theta}},\bm{\Sigma}_q\left(t\right)\right)}\\
&=\argmin_{\bm{\theta}}\frac{1}{2\sigma_q^2(t)}\left[\left\lVert\frac{1}{\sqrt{\alpha_t}}\vx_t + \frac{1 - \alpha_t}{\sqrt{\alpha_t}}\vs_{\bm{\theta}}(\vx_t, t) - 
\frac{1}{\sqrt{\alpha_t}}\vx_t - \frac{1 - \alpha_t}{\sqrt{\alpha_t}}\nabla\log p(\vx_t)\right\rVert_2^2\right]\\
&=\argmin_{\bm{\theta}}\frac{1}{2\sigma_q^2(t)}\left[\left\lVert \frac{1 - \alpha_t}{\sqrt{\alpha_t}}\vs_{\bm{\theta}}(\vx_t, t) - \frac{1 - \alpha_t}{\sqrt{\alpha_t}}\nabla\log p(\vx_t)\right\rVert_2^2\right]\\
&=\argmin_{\bm{\theta}}\frac{1}{2\sigma_q^2(t)}\left[\left\lVert \frac{1 - \alpha_t}{\sqrt{\alpha_t}}(\vs_{\bm{\theta}}(\vx_t, t) - \nabla\log p(\vx_t))\right\rVert_2^2\right]\\
&=\argmin_{\bm{\theta}}\frac{1}{2\sigma_q^2(t)}\frac{(1 - \alpha_t)^2}{\alpha_t}\left[\left\lVert \vs_{\bm{\theta}}(\vx_t, t) - \nabla\log p(\vx_t)\right\rVert_2^2\right] \label{eq:123}
\end{align}
Here, $\vs_{\bm{\theta}}(\vx_t, t)$ is a neural network that learns to predict the score function $\nabla_{\vx_t}\log p(\vx_t)$, which is the gradient of $\vx_t$ in data space, for any arbitrary noise level $t$.

The astute reader will notice that the score function $\nabla\log p(\vx_t)$ looks remarkably similar in form to the source noise $\bm{\epsilon}_0$.  This can be shown explicitly by combining Tweedie's Formula (Equation \ref{eq:109}) with the reparameterization trick (Equation \ref{eq:62}):
\begin{align}
\vx_0 = \frac{\vx_t + (1 - \bar\alpha_t)\nabla\log p(\vx_t)}{\sqrt{\bar\alpha_t}} &= \frac{\vx_t - \sqrt{1 - \bar\alpha_t}\bm{\epsilon}_0}{\sqrt{\bar\alpha_t}}\\
\therefore (1 - \bar\alpha_t)\nabla\log p(\vx_t) &= -\sqrt{1 - \bar\alpha_t}\bm{\epsilon}_0\\
\nabla\log p(\vx_t) &= -\frac{1}{\sqrt{1 - \bar\alpha_t}}\bm{\epsilon}_0
\end{align}
As it turns out, the two terms are off by a constant factor that scales with time!  The score function measures how to move in data space to maximize the log probability; intuitively, since the source noise is added to a natural image to corrupt it, moving in its opposite direction "denoises" the image and would be the best update to increase the subsequent log probability.  Our mathematical proof justifies this intuition; we have explicitly shown that learning to model the score function is equivalent to modeling the negative of the source noise (up to a scaling factor).

We have therefore derived three equivalent objectives to optimize a VDM: learning a neural network to predict the original image $\vx_0$, the source noise $\bm{\epsilon}_0$, or the score of the image at an arbitrary noise level $\nabla\log p(\vx_t)$.  The VDM can be scalably trained by stochastically sampling timesteps $t$ and minimizing the norm of the prediction with the ground truth target.

\section*{Score-based Generative Models}
\addcontentsline{toc}{section}{\protect\numberline{}Score-based Generative Models}%
We have shown that a Variational Diffusion Model can be learned simply by optimizing a neural network $\vs_{\bm{\theta}}(\vx_t, t)$ to predict the score function $\nabla\log p(\vx_t)$.  However, in our derivation, the score term arrived from an application of Tweedie's Formula; this doesn't necessarily provide us with great intuition or insight into what exactly the score function is or why it is worth modeling.  Fortunately, we can look to another class of generative models, Score-based Generative Models~\cite{song2019generative, song2020score, song2020improved}, for exactly this intuition.  As it turns out, we can show that the VDM formulation we have previously derived has an equivalent Score-based Generative Modeling formulation, allowing us to flexibly switch between these two interpretations at will.

To begin to understand why optimizing a score function makes sense, we take a detour and revisit energy-based models~\cite{lecun2006tutorial, song2021train}.  Arbitrarily flexible probability distributions can be written in the form:
\begin{align}
    p_{\bm{\theta}}(\vx) = \frac{1}{Z_{\bm{\theta}}}e^{-f_{\bm{\theta}}(\vx)} \label{eq:127}
\end{align}
where $f_{\bm{\theta}}(\vx)$ is an arbitrarily flexible, parameterizable function called the energy function, often modeled by a neural network, and $Z_{\bm{\theta}}$ is a normalizing constant to ensure that $\int p_{\bm{\theta}}(\vx)d\vx = 1$.  One way to learn such a distribution is maximum likelihood; however, this requires tractably computing the normalizing constant $Z_{\bm{\theta}} = \int e^{-f_{\bm{\theta}}(\vx)}d\vx$, which may not be possible for complex $f_{\bm{\theta}}(\vx)$ functions.

\begin{figure}
  \centering
  \includegraphics[width=0.45\linewidth]{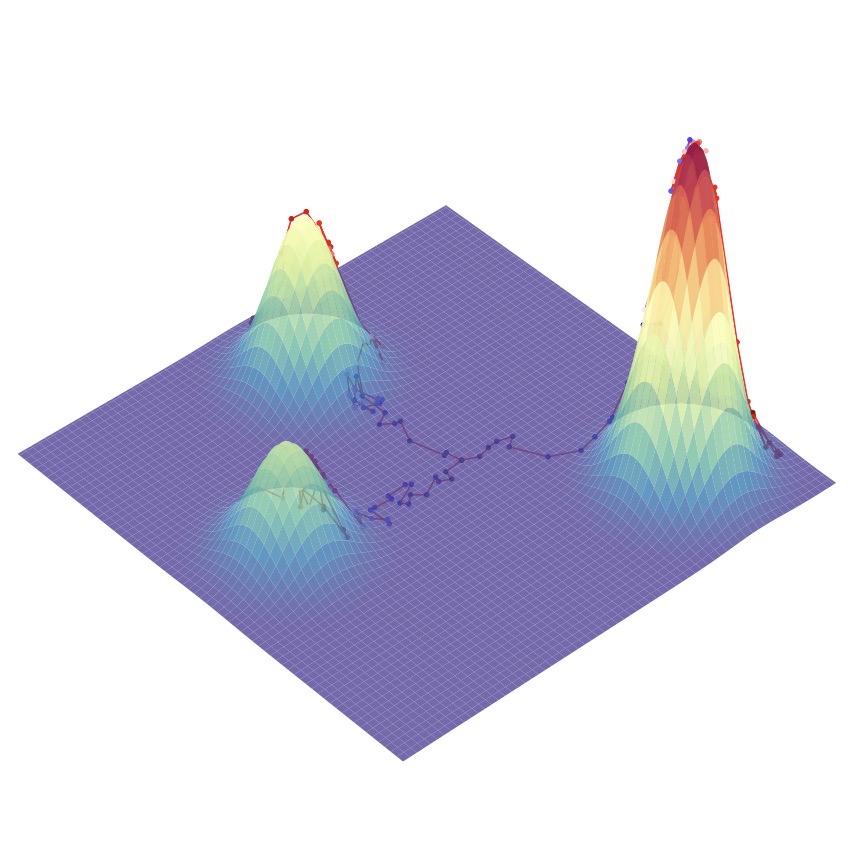}
  \includegraphics[width=0.4\linewidth]{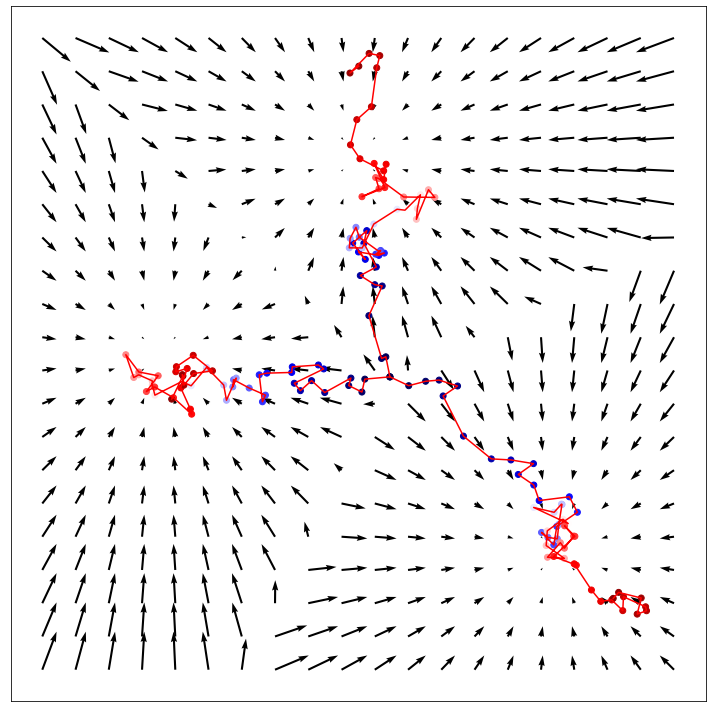}
  \caption{Visualization of three random sampling trajectories generated with Langevin dynamics, all starting from the same initialization point, for a Mixture of Gaussians.  The left figure plots these sampling trajectories on a three-dimensional contour, while the right figure plots the sampling trajectories against the ground-truth score function.  From the same initialization point, we are able to generate samples from different modes due to the stochastic noise term in the Langevin dynamics sampling procedure; without it, sampling from a fixed point would always deterministically follow the score to the same mode every trial.}
  \label{fig:langevin_sample}
\end{figure}

One way to avoid calculating or modeling the normalization constant is by using a neural network $\vs_{\bm{\theta}}(\vx)$ to learn the score function $\nabla\log p(\vx)$  of distribution  $p(\vx)$ instead.  This is motivated by the observation that taking the derivative of the log of both sides of Equation \ref{eq:127} yields:
\begin{align}
\nabla_\vx \log p_{\bm{\theta}}(\vx)
&= \nabla_\vx\log(\frac{1}{Z_{\bm{\theta}}}e^{-f_{\bm{\theta}}(\vx)})\\
&= \nabla_\vx\log\frac{1}{Z_{\bm{\theta}}} + \nabla_\vx\log e^{-f_{\bm{\theta}}(\vx)}\\
&= -\nabla_\vx f_{\bm{\theta}}(\vx)\\
&\approx \vs_{\bm{\theta}}(\vx)
\end{align}
which can be freely represented as a neural network without involving any normalization constants.  The score model can be optimized by minimizing the Fisher Divergence with the ground truth score function:
\begin{align}
    \mathbb{E}_{p(\vx)}\left[\left\lVert \vs_{\bm{\theta}}(\vx) - \nabla\log p(\vx)\right\rVert_2^2\right] \label{eq:132}
\end{align}
What does the score function represent?  For every $\vx$, taking the gradient of its log likelihood with respect to $\vx$ essentially describes what direction in data space to move in order to further increase its likelihood.  Intuitively, then, the score function defines a vector field over the entire space that data $\vx$ inhabits, pointing towards the modes.  Visually, this is depicted in the right plot of Figure \ref{fig:langevin_sample}.  Then, by learning the score function of the true data distribution, we can generate samples by starting at any arbitrary point in the same space and iteratively following the score until a mode is reached.  This sampling procedure is known as Langevin dynamics, and is mathematically described as:
\begin{align}
    \vx_{i+1} \leftarrow \vx_i + c\nabla\log p(\vx_i) + \sqrt{2c}\bm{\epsilon},\quad i = 0, 1, ..., K
\end{align}
where $\vx_0$ is randomly sampled from a prior distribution (such as uniform), and $\bm{\epsilon} \sim \mathcal{N}(\bm{\epsilon};\bm{0}, \textbf{I})$ is an extra noise term to ensure that the generated samples do not always collapse onto a mode, but hover around it for diversity.  Furthermore, because the learned score function is deterministic, sampling with a noise term involved adds stochasticity to the generative process, allowing us to avoid deterministic trajectories.  This is particularly useful when sampling is initialized from a position that lies between multiple modes.  A visual depiction of Langevin dynamics sampling and the benefits of the noise term is shown in Figure \ref{fig:langevin_sample}.

Note that the objective in Equation \ref{eq:132} relies on having access to the ground truth score function, which is unavailable to us for complex distributions such as the one modeling natural images.  Fortunately, alternative techniques known as score matching~\cite{hyvarinen2005estimation, saremi2018deep, song2020sliced, vincent2011connection} have been derived to minimize this Fisher divergence without knowing the ground truth score, and can be optimized with stochastic gradient descent.

Collectively, learning to represent a distribution as a score function and using it to generate samples through Markov Chain Monte Carlo techniques, such as Langevin dynamics, is known as Score-based Generative Modeling~\cite{song2019generative, song2020score, song2020improved}.

There are three main problems with vanilla score matching, as detailed by ~\citet{song2019generative}.  Firstly, the score function is ill-defined when $\vx$ lies on a low-dimensional manifold in a high-dimensional space.  This can be seen mathematically; all points not on the low-dimensional manifold would have probability zero, the log of which is undefined.  This is particularly inconvenient when trying to learn a generative model over natural images, which is known to lie on a low-dimensional manifold of the entire ambient space.

Secondly, the estimated score function trained via vanilla score matching will not be accurate in low density regions.  This is evident from the objective we minimize in Equation \ref{eq:132}.  Because it is an expectation over $p(\vx)$, and explicitly trained on samples from it, the model will not receive an accurate learning signal for rarely seen or unseen examples.  This is problematic, since our sampling strategy involves starting from a random location in the high-dimensional space, which is most likely random noise, and moving according to the learned score function.  Since we are following a noisy or inaccurate score estimate, the final generated samples may be suboptimal as well, or require many more iterations to converge on an accurate output.

Lastly, Langevin dynamics sampling may not mix, even if it is performed using the ground truth scores.  Suppose that the true data distribution is a mixture of two disjoint distributions:
\begin{align}
    p(\vx) = c_1p_1(\vx) + c_2p_2(\vx)
\end{align}
Then, when the score is computed, these mixing coefficients are lost, since the log operation splits the coefficient from the distribution and the gradient operation zeros it out.  To visualize this, note that the ground truth score function shown in the right Figure \ref{fig:langevin_sample} is agnostic of the different weights between the three distributions; Langevin dynamics sampling from the depicted initialization point has a roughly equal chance of arriving at each mode, despite the bottom right mode having a higher weight in the actual Mixture of Gaussians.

It turns out that these three drawbacks can be simultaneously addressed by adding multiple levels of Gaussian noise to the data.  Firstly, as the support of a Gaussian noise distribution is the entire space, a perturbed data sample will no longer be confined to a low-dimensional manifold.  Secondly, adding large Gaussian noise will increase the area each mode covers in the data distribution, adding more training signal in low density regions.  Lastly, adding multiple levels of Gaussian noise with increasing variance will result in intermediate distributions that respect the ground truth mixing coefficients.

\newpage
Formally, we can choose a positive sequence of noise levels $\{\sigma_t\}_{t=1}^T$ and define a sequence of progressively perturbed data distributions:
\begin{align}
p_{\sigma_t}(\vx_t) = \int p(\vx)\mathcal{N}(\vx_t; \vx, \sigma_t^2\textbf{I})d\vx
\end{align}
Then, a neural network $\vs_{\bm{\theta}}(\vx, t)$ is learned using score matching to learn the score function for all noise levels simultaneously:
\begin{align}
\argmin_{\bm{\theta}} \sum_{t=1}^T\lambda(t)\mathbb{E}_{p_{\sigma_t}(\vx_t)}\left[\left\lVert \vs_{\bm{\theta}}(\vx, t) - \nabla\log p_{\sigma_t}(\vx_t)\right\rVert_2^2\right]
\end{align}
where $\lambda(t)$ is a positive weighting function that conditions on noise level $t$.  Note that this objective almost exactly matches the objective derived in Equation \ref{eq:123} to train a Variational Diffusion Model.  Furthermore, the authors propose annealed Langevin dynamics sampling as a generative procedure, in which samples are produced by running Langevin dynamics for each $t = T, T-1, ..., 2, 1$ in sequence.  The initialization is chosen from some fixed prior (such as uniform), and each subsequent sampling step starts from the final samples of the previous simulation.  Because the noise levels steadily decrease over timesteps $t$, and we reduce the step size over time, the samples eventually converge into a true mode.  This is directly analogous to the sampling procedure performed in the Markovian HVAE interpretation of a Variational Diffusion Model, where a randomly initialized data vector is iteratively refined over decreasing noise levels.

Therefore, we have established an explicit connection between Variational Diffusion Models and Score-based Generative Models, both in their training objectives and sampling procedures.

One question is how to naturally generalize diffusion models to an infinite number of timesteps.  Under the Markovian HVAE view, this can be interpreted as extending the number of hierarchies to infinity $T \rightarrow \infty$.  It is clearer to represent this from the equivalent score-based generative model perspective; under an infinite number of noise scales, the perturbation of an image over continuous time can be represented as a stochastic process, and therefore described by a stochastic differential equation (SDE).  Sampling is then performed by reversing the SDE, which naturally requires estimating the score function at each continuous-valued noise level~\cite{song2020score}.  Different parameterizations of the SDE essentially describe different perturbation schemes over time, enabling flexible modeling of the noising procedure~\cite{kingma2021variational}.

\section*{Guidance}
\addcontentsline{toc}{section}{\protect\numberline{}Guidance}%
So far, we have focused on modeling just the data distribution $p(\vx)$.  However, we are often also interested in learning conditional distribution $p(\vx|y)$, which would enable us to explicitly control the data we generate through conditioning information $y$.  This forms the backbone of image super-resolution models such as Cascaded Diffusion Models~\cite{ho2022cascaded}, as well as state-of-the-art image-text models such as DALL-E 2~\cite{ramesh2022hierarchical} and Imagen~\cite{saharia2022photorealistic}.

A natural way to add conditioning information is simply alongside the timestep information, at each iteration.  Recall our joint distribution from Equation \ref{eq:36}:
$$p(\vx_{0:T}) = p(\vx_T)\prod_{t=1}^Tp_{\bm{\theta}}(\vx_{t-1}|\vx_t)$$
Then, to turn this into a conditional diffusion model, we can simply add arbitrary conditioning information $y$ at each transition step as:
\begin{align}
p(\vx_{0:T}|y) = p(\vx_T)\prod_{t=1}^Tp_{\bm{\theta}}(\vx_{t-1}|\vx_t, y)
\end{align}
For example, $y$ could be a text encoding in image-text generation, or a low-resolution image to perform super-resolution on.  We are thus able to learn the core neural networks of a VDM as before, by predicting $\hat \vx_{\bm{\theta}}(\vx_t, t, y) \approx \vx_0$, $\bm{\hat\epsilon}_{\bm{\theta}}(\vx_t, t, y) \approx \bm{\epsilon}_0$, or $\vs_{\bm{\theta}}(\vx_t, t, y) \approx \nabla\log p(\vx_t|y)$ for each desired interpretation and implementation.

A caveat of this vanilla formulation is that a conditional diffusion model trained in this way may potentially learn to ignore or downplay any given conditioning information.  Guidance is therefore proposed as a way to more explicitly control the amount of weight the model gives to the conditioning information, at the cost of sample diversity.  The two most popular forms of guidance are known as Classifier Guidance~\cite{song2020score, dhariwal2021diffusion} and Classifier-Free Guidance~\cite{ho2021classifier}.

\subsubsection*{Classifier Guidance}
\addcontentsline{toc}{section}{\protect\numberline{}\protect\numberline{}Classifier Guidance}%
Let us begin with the score-based formulation of a diffusion model, where our goal is to learn $\nabla\log p(\vx_t|y)$, the score of the conditional model, at arbitrary noise levels $t$.  Recall that $\nabla$ is shorthand for $\nabla_{\vx_t}$ in the interest of brevity.  By Bayes rule, we can derive the following equivalent form:
\begin{align}
\nabla\log p(\vx_t|y) &= \nabla\log \left( \frac{p(\vx_t)p(y|\vx_t)}{p(y)} \right)\\
&= \nabla\log p(\vx_t) + \nabla\log p(y|\vx_t) - \nabla\log p(y)\\
&= \underbrace{\nabla\log p(\vx_t)}_\text{unconditional score} + \underbrace{\nabla\log p(y|\vx_t)}_\text{adversarial gradient} \label{eq:148}
\end{align}
where we have leveraged the fact that the gradient of $\log p(y)$ with respect to $\vx_t$ is zero.

Our final derived result can be interpreted as learning an unconditional score function combined with the adversarial gradient of a classifier $p(y|\vx_t)$.  Therefore, in Classifier Guidance~\cite{song2020score, dhariwal2021diffusion}, the score of an unconditional diffusion model is learned as previously derived, alongside a classifier that takes in arbitrary noisy $\vx_t$ and attempts to predict conditional information $y$.  Then, during the sampling procedure, the overall conditional score function used for annealed Langevin dynamics is computed as the sum of the unconditional score function and the adversarial gradient of the noisy classifier.

In order to introduce fine-grained control to either encourage or discourage the model to consider the conditioning information, Classifier Guidance scales the adversarial gradient of the noisy classifier by a $\gamma$ hyperparameter term.  The score function learned under Classifier Guidance can then be summarized as:
\begin{align}
    \nabla\log p(\vx_t|y) &= \nabla\log p(\vx_t) + \gamma\nabla\log p(y|\vx_t) \label{eq:150}
\end{align}
Intuitively, when $\gamma=0$ the conditional diffusion model learns to ignore the conditioning information entirely, and when $\gamma$ is large the conditional diffusion model learns to produce samples that heavily adhere to the conditioning information.  This would come at the cost of sample diversity, as it would only produce data that would be easy to regenerate the provided conditioning information from, even at noisy levels.

One noted drawback of Classifier Guidance is its reliance on a separately learned classifier.  Because the classifier must handle arbitrarily noisy inputs, which most existing pretrained classification models are not optimized to do, it must be learned ad hoc alongside the diffusion model.

\subsubsection*{Classifier-Free Guidance}
\addcontentsline{toc}{section}{\protect\numberline{}\protect\numberline{}Classifier-Free Guidance}%
In Classifier-Free Guidance~\cite{ho2021classifier}, the authors ditch the training of a separate classifier model in favor of an unconditional diffusion model and a conditional diffusion model.  To derive the score function under Classifier-Free Guidance, we can first rearrange Equation \ref{eq:148} to show that:
\begin{align}
    \nabla\log p(y|\vx_t) = \nabla\log p(\vx_t|y) - \nabla\log p(\vx_t)
\end{align}
Then, substituting this into Equation \ref{eq:150}, we get:
\begin{align}
\nabla\log p(\vx_t|y)
&= \nabla\log p(\vx_t) + \gamma\left(\nabla\log p(\vx_t|y) - \nabla\log p(\vx_t)\right)\\
&= \nabla\log p(\vx_t) + \gamma\nabla\log p(\vx_t|y) - \gamma\nabla\log p(\vx_t)\\
&= \underbrace{\gamma\nabla\log p(\vx_t|y)}_\text{conditional score} + \underbrace{(1 - \gamma)\nabla\log p(\vx_t)}_\text{unconditional score}
\end{align}
Once again, $\gamma$ is a term that controls how much our learned conditional model cares about the conditioning information.  When $\gamma = 0$, the learned conditional model completely ignores the conditioner and learns an unconditional diffusion model.  When $\gamma = 1$, the model explicitly learns the vanilla conditional distribution without guidance.  When $\gamma > 1$, the diffusion model not only prioritizes the conditional score function, but also moves in the direction away from the unconditional score function.  In other words, it reduces the probability of generating samples that do not use conditioning information, in favor of the samples that explicitly do.  This also has the effect of decreasing sample diversity at the cost of generating samples that accurately match the conditioning information.

Because learning two separate diffusion models is expensive, we can learn both the conditional and unconditional diffusion models together as a singular conditional model; the unconditional diffusion model can be queried by replacing the conditioning information with fixed constant values, such as zeros.  This is essentially performing random dropout on the conditioning information.  Classifier-Free Guidance is elegant because it enables us greater control over our conditional generation procedure while requiring nothing beyond the training of a singular diffusion model.

\section*{Closing}
\addcontentsline{toc}{section}{\protect\numberline{}Closing}%
Allow us to recapitulate our findings over the course of our explorations.  First, we derive Variational Diffusion Models as a special case of a Markovian Hierarchical Variational Autoencoder, where three key assumptions enable tractable computation and scalable optimization of the ELBO.  We then prove that optimizing a VDM boils down to learning a neural network to predict one of three potential objectives: the original source image from any arbitrary noisification of it, the original source noise from any arbitrarily noisified image, or the score function of a noisified image at any arbitrary noise level.  Then, we dive deeper into what it means to learn the score function, and connect it explicitly with the perspective of Score-based Generative Modeling.  Lastly, we cover how to learn a conditional distribution using diffusion models.

In summary, diffusion models have shown incredible capabilities as generative models; indeed, they power the current state-of-the-art models on text-conditioned image generation such as Imagen and DALL-E 2.  Furthermore, the mathematics that enable these models are exceedingly elegant.  However, there still remain a few drawbacks to consider:
\begin{itemize}
    \item It is unlikely that this is how we, as humans, naturally model and generate data; we do not generate samples as random noise that we iteratively denoise.
    \item The VDM does not produce interpretable latents.  Whereas a VAE would hopefully learn a structured latent space through the optimization of its encoder, in a VDM the encoder at each timestep is already given as a linear Gaussian model and cannot be optimized flexibly.  Therefore, the intermediate latents are restricted as just noisy versions of the original input.
    \item The latents are restricted to the same dimensionality as the original input, further frustrating efforts to learn meaningful, compressed latent structure.
    \item Sampling is an expensive procedure, as multiple denoising steps must be run under both formulations.  Recall that one of the restrictions is that a large enough number of timesteps $T$ is chosen to ensure the final latent is completely Gaussian noise; during sampling we must iterate over all these timesteps to generate a sample.
\end{itemize}

As a final note, the success of diffusion models highlights the power of Hierarchical VAEs as a generative model.  We have shown that when we generalize to \textit{infinite} latent hierarchies, even if the encoder is trivial and the latent dimension is fixed and Markovian transitions are assumed, we are still able to learn powerful models of data.  This suggests that further performance gains can be achieved in the case of general, deep HVAEs, where complex encoders and semantically meaningful latent spaces can be potentially learned.

\textbf{Acknowledgments:} I would like to acknowledge Josh Dillon, Yang Song, Durk Kingma, Ben Poole, Jonathan Ho, Yiding Jiang, Ting Chen, Jeremy Cohen, and Chen Sun for reviewing drafts of this work and providing many helpful edits and comments.  Thanks so much!

\newpage
{\small
\bibliography{ref}
\bibliographystyle{unsrtnat}}

\end{document}